\documentclass{article}

\usepackage{graphicx}
\usepackage{amsmath}
\usepackage{amssymb}
\usepackage{booktabs}
\usepackage{multirow}
\usepackage{wrapfig}
\usepackage{longtable}
\usepackage{booktabs}
\usepackage[most]{tcolorbox}
\usepackage[table]{xcolor}
\usepackage{subcaption}

\newcommand\blfootnote[1]{%
  \begingroup
  \renewcommand\thefootnote{}\footnote{#1}%
  \addtocounter{footnote}{-1}%
  \endgroup}
\usepackage[preprint]{corl_2026} 

\title{Direct Action-Head Injection of A Grounded 3D Point Unlocks Spatial and Task Generalization}

\author{
    \textbf{Shiang-Feng Tsai}$^{1}$ \quad \textbf{Jin-Cheng Jhang}$^{1}$ \quad \textbf{Yen-Ling Tai}$^{2}$ \\[2pt] \quad \textbf{Jia-Hong Lai}$^{*1}$ \quad \textbf{Shih-Yun Wong}$^{*1}$ \quad \textbf{Kang-Tung Hsu}$^{*1}$ \quad \textbf{Yi-Ting Chen}$^{2}$ \quad \textbf{Min Sun}$^{1}$ \\[2pt]
    $^{1}$National Tsing Hua University \quad $^{2}$National Yang Ming Chiao Tung University
}

\begin{document}
\maketitle
\blfootnote{$^{*}$Equal contribution.}

\begin{abstract}
Vision-Language-Action (VLA) models leverage large-scale vision-language pretraining for flexible robot manipulation, yet at test time they remain brittle to changed object positions and to familiar scenes paired with different instructions. 
A growing family of methods addresses this brittleness by supplying the policy with grounding signals, such as 2D pixel coordinates for object localization and placement. 
However, we find that how the grounding signal is represented and injected matters more than the signal itself. In this work, we propose a lightweight module that represents the grounding signal in 3D and injects the resulting embedding directly into the action head. The module is a two-layer MLP and requires no changes to the VLA backbone or pretraining pipeline, yet it yields substantially larger gains than language- or visual-prompting alternatives. 
On LIBERO-PRO, our method improves the average success rate of GR00T-N1.6 from $31.2$ to $77.5$ under task perturbation and from $28.1$ to $60.2$ under position perturbation. Comparable gains are also achieved for $\pi_{0.5}$, demonstrating that the mechanism is backbone-agnostic across VLAs with diffusion-based action heads. 
We further validate the practical applicability with real-world experiments. 
Together, these results support our central finding: lifting adequate 2D grounding into 3D and injecting it into the action head enables spatial and instance-level task generalization in VLAs. 
Project page: \href{https://3dgnd-head-injection.github.io/}{\texttt{3dgnd-head-injection.github.io}}

\end{abstract}


\section{Introduction}
\label{sec:introduction}
Vision-Language-Action (VLA) models~\cite{rt22023arxiv, octo_2023, kim2024openvla, pi0, pi05, gr00tn1_2025} inherit rich semantic representations from large-scale vision-language pretraining, enabling them to follow free-form instructions and recognize a wide range of objects during manipulation. However, VLAs remain brittle in test-time generalization along two distinct axes—\emph{spatial generalization}, where object positions differ from those seen during training~\cite{zhou2025liberopro, fei25libero-plus}, and instance-level \emph{task generalization}, where a familiar scene is paired with an instruction specifying a different target at test time~\cite{zhou2025liberopro}. Closing these generalization gaps has become a central challenge for the field.

A growing line of work addresses this by providing the policy with explicit spatial grounding---the location of the task-relevant object or placement region. One family relies on external grounding modules such as Vision-Language Models (VLMs) or detectors~\cite{yuan2024robopoint, roboground} and conveys their predictions to the policy as coordinates in the language prompt~\cite{molmobot}, visual prompt~\cite{moka, wang2026vp}, or combined visual--textual conditioning~\cite{coavla}. The other trains the VLA itself to emit such spatial outputs as an intermediate chain-of-thought prediction before producing actions~\cite{molmoact2025, ecot, steerablepolicy}. While both directions improve generalization, neither systematically addresses how the spatial signal should be represented and injected into the policy to achieve robust generalization.

\begin{wrapfigure}{r}{0.45\textwidth}
    \includegraphics[width=\linewidth]{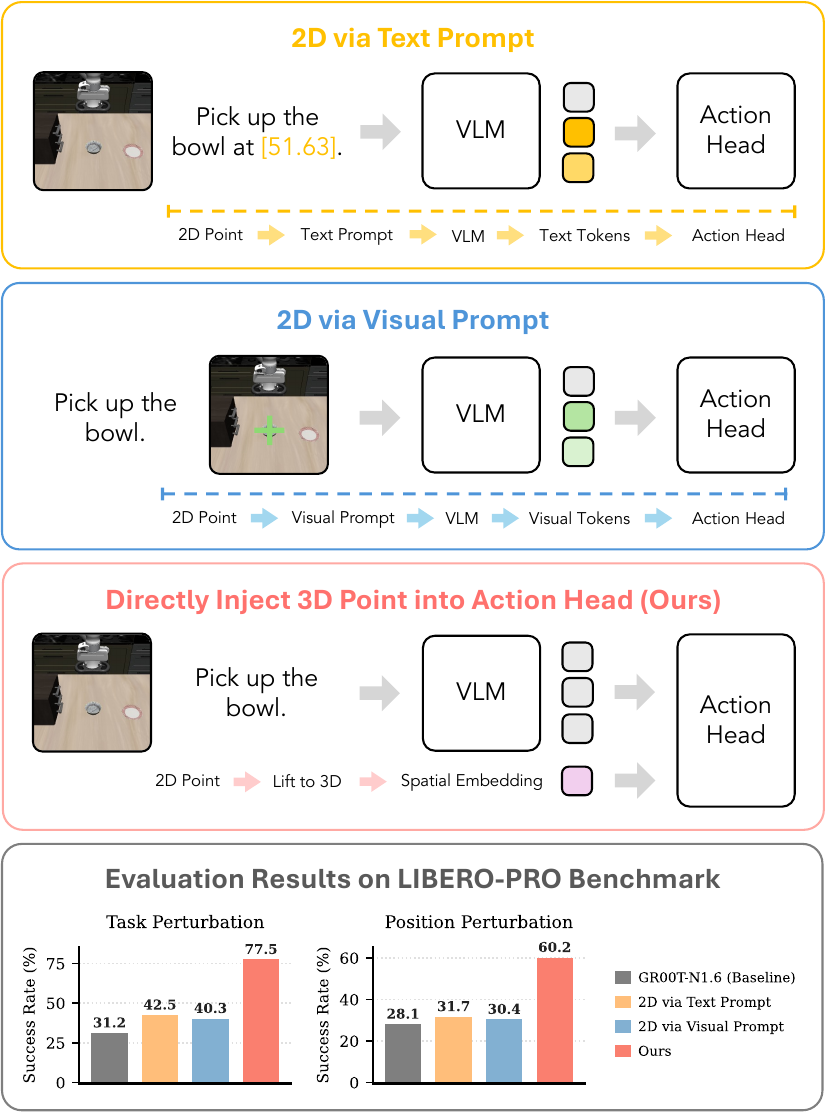}
    \caption{\textbf{Comparison of grounding representation and injection paradigms.} Given an identical 2D target point, appending coordinates to the language prompt and overlaying it as a visual prompt on the image yield slight improvements over the baseline, whereas lifting it into 3D and injecting it into the action head leads to substantially better performance on LIBERO-PRO benchmark.} 
    \label{fig:teaser}
\end{wrapfigure}

Specifically, as shown in Fig.~\ref{fig:teaser}, we show that lifting a 2D point through depth into 3D and injecting it directly into the action head as a spatial embedding is more effective at unlocking robust generalization than relaying it through 2D proxies such as language tokens or image overlays. 
The reason is intuitive: action prediction operates in the 3D space, so a 3D signal matches the geometry of the output, whereas 2D representations force the policy to internally reconstruct this geometry from visual and linguistic features. The same mechanism also explains improvements on task generalization: an explicit spatial signal tied to the current instruction prevents the policy from defaulting to scene-triggered trajectories memorized during training, forcing it to act on the target specified by the instruction at test time.


To realize this design, we propose a lightweight, model-agnostic module that lifts the grounding signal into 3D and feeds it into the action head. Specifically, we lift the 2D target point into 3D using depth and camera parameters and compute its displacement relative to the gripper. We then use a two-layer MLP to encode this displacement into a spatial embedding. Modern VLAs commonly adopt a diffusion transformer (DiT) as the action head~\cite{li2024cogact, pi0, pi05, gr00tn1_2025, molmobot}; we add the spatial embedding to the existing time-step embedding, and the combined embedding conditions the DiT via adaptive layer normalization~\cite{DiT, perez2018film}. The module requires no changes to the VLA backbone or pretraining pipeline; the only added component is a two-layer MLP.

We validate this design through extensive simulation experiments on LIBERO~\cite{libero} and LIBERO-PRO~\cite{zhou2025liberopro}, complemented by real-world evaluations. Applied on top of GR00T-N1.6, our module raises the average LIBERO-PRO success rate from $31.2\%$ to \textbf{77.5}$\%$ under Task Perturbation and from $28.1\%$ to \textbf{60.2}$\%$ under Position Perturbation. Our method yields comparable gains when applied to $\pi_{0.5}$ ($37.3\%$ to \textbf{75.9}$\%$ and $47.0\%$ to \textbf{72.2}$\%$ under the two perturbations), demonstrating that the mechanism is backbone-agnostic across VLAs with diffusion-based action heads. 
We further validate our method in the real world, where it remains effective under both perturbations and substantially outperforms the baselines.
These results point to a simple but overlooked finding: with adequate 2D grounding in hand, generalization hinges on how the signal is represented and injected—lifting it into 3D and injecting it directly into the action head substantially improves both spatial and instance-level task generalization.
\section{Related Work}
\noindent\textbf{Vision-Language-Action Models.}
Vision-Language-Action (VLA) models build upon pretrained Vision-Language Models (VLMs) to enable language-conditioned robot control. Early work such as RT-2~\cite{rt22023arxiv} fine-tunes a VLM to output discretized action tokens. Subsequent efforts scale this paradigm through larger datasets and diverse embodiments~\cite{open_x_embodiment_rt_x_2023, octo_2023, khazatsky2024droid, kim2024openvla}, and more recent models have converged on a common architectural pattern: pairing a VLM backbone with a diffusion or flow-matching action head~\cite{li2024cogact, pi0, pi05, gr00tn1_2025, molmobot}. Despite rapid progress, recent stress tests reveal that these models remain brittle along two complementary axes: \emph{spatial generalization}, where success rates collapse under changes to object positions~\cite{zhou2025liberopro, fei25libero-plus}, and \emph{task generalization}, where models replay memorized trajectories rather than respond to changes in the instruction~\cite{zhou2025liberopro}. Closing these gaps has emerged as a central challenge in the development of robust VLAs.

\noindent\textbf{Spatial Grounding for Vision-Language-Action Models.}
A growing line of work addresses the mentioned generalization gaps by endowing the policy with explicit grounding signals indicating where the task-relevant object or region lies. One family relies on external grounding modules---typically VLMs or detectors that predict pixel coordinates~\cite{yuan2024robopoint}, 2D trajectories~\cite{li2025hamster}, or segmentation masks~\cite{roboground}---and convey their predictions to the policy either through language prompt~\cite{molmobot} or visual prompt~\cite{moka, wang2026vp, li2025hamster}. Another family instead trains the VLA itself to emit such spatial outputs as an intermediate chain-of-thought prediction before producing actions~\cite{coavla, molmoact2025, ecot, steerablepolicy}, with some further routing these predictions through dedicated injection modules to the action head~\cite{coavla}. Despite this diversity in source and routing, the spatial signal in prior work remains a 2D quantity tied to the image plane, and how this signal is represented and injected into the policy has received little systematic attention. We show that this is the true bottleneck, and that a simple change to both---lifting the signal into 3D and injecting it directly into the action head---unlocks substantial gains.

\noindent\textbf{3D-Enhanced Vision-Language-Action Models.}
The value of 3D information for manipulation has been demonstrated by a line of work that introduces explicit 3D representations into VLAs, ranging from point cloud alignment in 3D-based LLMs~\cite{zhen20243dvla} and ego-centric position encodings in the backbone~\cite{spatialvla} to dense spatial tokens injected into the action head~\cite{pointvla, falcon}. These methods consistently show that richer 3D information improves manipulation, but obtain it through dense scene-level representations that require a dedicated 3D encoder---and in some cases an additional foundation model---on top of the VLA. Our work pushes this observation to its minimal form: starting from a 2D grounding signal already available from off-the-shelf detectors or VLMs, a single task-relevant point lifted into 3D---encoding only the gripper-to-target displacement---already delivers substantial gains, with no additional 3D encoder and no changes to the backbone.
\section{Method}
\begin{figure}[t]
\centering
\includegraphics[width=\linewidth]{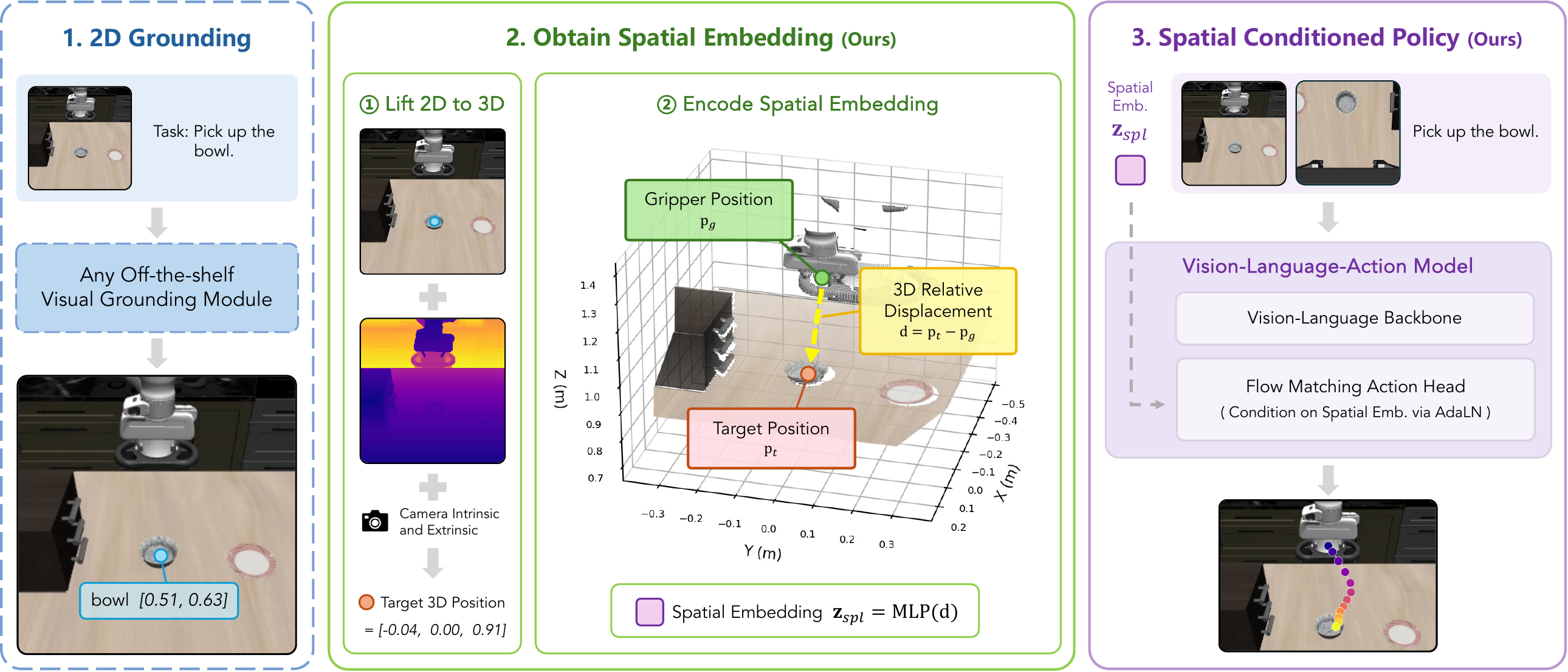}
\caption{\textbf{Overview of our proposed pipeline.} Given a task instruction, (1) any off-the-shelf visual grounding module localizes the task-relevant object as a 2D point in the image. (2) We lift this 2D point into 3D, compute its relative displacement to the gripper, and encode the displacement through a two-layer MLP into a spatial embedding. (3) The spatial embedding is injected directly into the flow-matching action head via AdaLN conditioning.}
\label{fig:overall_pipeline}
\end{figure}
We instantiate the two design choices: representing the grounding signal in 3D, and injecting it directly into the action head. Given a 2D target point—obtainable from any off-the-shelf grounding source—together with depth and camera parameters, we lift the point into 3D, encode the result into a spatial embedding via a two-layer MLP, and condition the action head on this embedding, with no changes to the VLA backbone or training objective. Fig.~\ref{fig:overall_pipeline} illustrates the overall pipeline.
 
\subsection{From 2D Target Point to 3D Spatial Embedding}
Our method is applicable to tasks whose sub-goals can each be associated with a target object or region. In pick-and-place tasks, for instance, the reach-to-grasp sub-goal corresponds to the target object and the transport-to-place sub-goal to the placement region. Given a 2D target point indicating the current sub-goal, we lift it into 3D using the depth image and camera parameters to obtain the target position $\mathbf{p}_t \in \mathbb{R}^3$ in the robot base frame. We also obtain the current gripper position $\mathbf{p}_g \in \mathbb{R}^3$ from the robot state. We then compute their relative displacement:
\begin{equation}
    \mathbf{d} = \mathbf{p}_t - \mathbf{p}_g
\end{equation}
We encode this displacement into a spatial embedding via a two-layer MLP, whose output dimensionality matches the hidden size of the action head:
\begin{equation}
    \mathbf{z}_{\text{spatial}} = \operatorname{MLP}(\mathbf{d}).
\end{equation}
Despite its simplicity, $\mathbf{z}_\text{spatial}$ directly captures the task-relevant 3D geometric relationship between the target and the gripper---the information most pertinent to action prediction.
 
\subsection{Inject Spatial Embedding into the VLA}
 
Prior works~\cite{pointvla, falcon} have empirically shown that injecting 3D information exclusively into the action head preserves the pretrained VLM backbone features while improving spatial performance. These methods introduce auxiliary 3D encoders and projection modules whose outputs are added directly to the action tokens. Building on this insight, we likewise inject the 3D signal into the action head, but instead route it through the existing adaptive layer normalization (AdaLN) conditioning mechanism~\cite{DiT, perez2018film} already present in the Diffusion Transformer~\cite{DiT} (DiT) action head~\cite{li2024cogact, pi0, pi05, gr00tn1_2025, molmobot}, requiring no changes to the backbone. In the original AdaLN formulation, conditioning is performed solely on the timestep embedding $\mathbf{z}_\text{time}$. We extend this by combining $\mathbf{z}_\text{time}$ and $\mathbf{z}_\text{spatial}$ via element-wise addition before regressing the scale and shift parameters $\gamma$ and $\beta$:
\begin{equation}
    \gamma, \beta = \operatorname{Linear}(\mathbf{z}_\text{time} + \mathbf{z}_\text{spatial}).
\end{equation}
These parameters modulate the normalized features in each DiT block of the action head, allowing the policy to incorporate task-relevant 3D geometric information when predicting actions. 

\subsection{Training and Inference}
During both training and inference, we use a fixed off-the-shelf grounding source to provide the grounding information. The grounding signal is used solely as input to the MLP to obtain the spatial embedding $\mathbf{z}_\text{spatial}$. The proposed MLP is trained jointly and end-to-end with the VLA using the original flow-matching action loss, with no additional supervision or auxiliary objective.
\section{Simulation Experiments}
\label{sec:sim_exp}
\subsection{Simulation Experimental Setup}
\noindent\textbf{Benchmarks.} We evaluate on LIBERO~\cite{libero} and LIBERO-PRO~\cite{zhou2025liberopro}. From LIBERO, we adopt two pick-and-place suites, LIBERO-Object and LIBERO-Spatial. LIBERO-PRO extends LIBERO with two test-time perturbations: Task Perturbation replaces the target object with a plausible alternative present in the scene, testing whether the policy follows the test-time instruction rather than replaying memorized task–trajectory associations; Position Perturbation alters initial object placements to evaluate spatial generalization. We run 50 rollout trials per task, totaling 500 trials per suite, and report average success rates.

\noindent\textbf{Training Data.}
We augment LIBERO-Object and LIBERO-Spatial with 49 additional pick-and-place tasks from LIBERO-90, yielding 69 tasks in total. The original LIBERO suites provide unimodal trajectories~\cite{zhou2025liberopro} and minimally randomized initial states, under which baselines trivially fail to generalize~\cite{steerablepolicy}. Scaling to 69 tasks increases diversity and ensures that any observed OOD collapse reflects a structural limitation rather than a data-side shortcoming. In Sec.~\ref{sec:tasks_scaling}, we also evaluate models trained on the original 20-task setup. The 49 tasks from LIBERO-90 are listed in App.~\ref{supp:sec:subsec:training_tasks}.

\noindent\textbf{Grounding Source and Sub-Goal Switching.}
We obtain 2D target points by sampling from simulator-provided segmentation masks of the target object or placement region. Both during training and at inference, the 3D target point is obtained by lifting a 2D pixel from the third-person view image. For more details please refer to App.~\ref{supp:sec:subsec:grounding-source}. We decompose each pick-and-place task into reach-to-grasp and transport-to-place sub-goals. Inspired by prior work~\cite{deng2025graspvla}, we record the target object's z-coordinate in 3D space at episode start and consider the object as grasped once it rises by more than 1 cm. Upon grasping, we switch the active target from the object to the placement region.

\noindent\textbf{Base Models.}
For the main result in Sec.~\ref{sec:libero_and_libero_pro}, we use two pre-trained VLA backbones:  $\pi_{0.5}$~\cite{pi05} and GR00T-N1.6~\cite{gr00tn1_2025}. For each backbone we fine-tune two variants on our 69-task suite that differ only in whether our method is applied. All subsequent experiments are conducted on GR00T-N1.6. Full training details are provided in App.~\ref{supp:sec:subsec:sim-training-details}.


\subsection{Experimental Results on LIBERO and LIBERO-PRO}
\label{sec:libero_and_libero_pro}
\definecolor{mygray}{HTML}{EFEFEF}
\definecolor{myblue}{HTML}{EAF3FF}
\definecolor{deltagreen}{HTML}{3A7D44}
\definecolor{deltared}{HTML}{CC3333}

\begin{table}[t]
\centering
\caption{\textbf{LIBERO and LIBERO-PRO results.} All baselines exhibit substantial performance drops under both Task Perturbation and Position Perturbation. In contrast, our method consistently improves two distinct backbones under both perturbations while preserving in-distribution performance, demonstrating that the mechanism is both effective and backbone-agnostic. The best performance in each suite is highlighted in bold. * denotes results reproduced by us using the same datasets as ours.}
\label{tab:libero_and_libero_pro}
\resizebox{0.9\linewidth}{!}{
\begin{tabular}{l ccc ccc ccc}
\toprule
\multirow{3}{*}{\textbf{Method}} 
  & \multicolumn{3}{c}{\textbf{LIBERO}~\cite{libero}} 
  & \multicolumn{6}{c}{\textbf{LIBERO-PRO}~\cite{zhou2025liberopro}} \\
\cmidrule(lr){2-4} \cmidrule(lr){5-10}
  & \multicolumn{3}{c}{In Distribution}
  & \multicolumn{3}{c}{Task Perturbation} 
  & \multicolumn{3}{c}{Position Perturbation} \\
\cmidrule(lr){2-4} \cmidrule(lr){5-7} \cmidrule(lr){8-10}
  & Object & Spatial & Avg. & Object & Spatial & Avg. & Object & Spatial & Avg. \\
\midrule

\rowcolor{mygray} \multicolumn{10}{l}{\textit{Open-source VLAs}} \\
\midrule
OpenVLA~\cite{kim2024openvla} & 88.4 & 84.7 & 86.6 & 1.8 & 46.4 & 24.1 & 0.0 & 0.0 & 0.0 \\
$\pi_{0}$~\cite{pi0} & 98.8 & 96.8 & 97.8 & 10.0 & 54.4 & 32.2 & 0.2 & 11.6 & 5.9 \\
MolmoAct~\cite{molmoact2025} & 95.4 & 87.0 & 91.2 & 9.2 & 31.6 & 20.4 & 4.2 & 0.2 & 2.2 \\
FALCON~\cite{falcon}* & 98.8 & 97.6 & \textbf{98.2} & 5.0 & 48.2 & 26.6 & 0.6 & 12.2 & 6.4 \\
\midrule

\rowcolor{mygray} \multicolumn{10}{l}{\textit{Controlled comparison}} \\
\midrule
$\pi_{0.5}$~\cite{pi05} & 94.8 & 96.0 & 95.4 & 20.0 & 54.6 & 37.3 & 36.0 & 58.0 & 47.0 \\
\rowcolor{myblue}
$\pi_{0.5}$ + Ours & 97.0 & \textbf{98.2} & 97.6 & \textbf{70.6} & 81.2 & 75.9 & 75.0 & \textbf{69.4} & \textbf{72.2} \\
\rowcolor{myblue}
$\Delta$ & \textcolor{deltagreen}{\footnotesize{2.2 $\uparrow$}} & \textcolor{deltagreen}{\footnotesize{2.2 $\uparrow$}} & \textcolor{deltagreen}{\footnotesize{2.2 $\uparrow$}} & \textcolor{deltagreen}{\footnotesize{50.6 $\uparrow$}} & \textcolor{deltagreen}{\footnotesize{26.6 $\uparrow$}} & \textcolor{deltagreen}{\footnotesize{38.6 $\uparrow$}} & \textcolor{deltagreen}{\footnotesize{39.0 $\uparrow$}} & \textcolor{deltagreen}{\footnotesize{11.4 $\uparrow$}} & \textcolor{deltagreen}{\footnotesize{25.2 $\uparrow$}} \\
\midrule
GR00T-N1.6~\cite{gr00tn1_2025} & 93.8 & 93.2 & 93.5 & 10.0 & 52.4 & 31.2 & 29.2 & 27.0 & 28.1 \\
\rowcolor{myblue}
GR00T-N1.6 + Ours & \textbf{99.6} & 92.4 & 96.0 & 70.4 & \textbf{84.6} & \textbf{77.5} & \textbf{79.4} & 41.0 & 60.2 \\
\rowcolor{myblue}
$\Delta$ & \textcolor{deltagreen}{\footnotesize{5.8 $\uparrow$}} & \textcolor{deltared}{\footnotesize{0.8 $\downarrow$}} & \textcolor{deltagreen}{\footnotesize{2.5 $\uparrow$}} & \textcolor{deltagreen}{\footnotesize{60.4 $\uparrow$}} & \textcolor{deltagreen}{\footnotesize{32.2 $\uparrow$}} & \textcolor{deltagreen}{\footnotesize{46.3 $\uparrow$}} & \textcolor{deltagreen}{\footnotesize{50.2 $\uparrow$}} & \textcolor{deltagreen}{\footnotesize{14.0 $\uparrow$}} & \textcolor{deltagreen}{\footnotesize{32.1 $\uparrow$}} \\
\bottomrule
\end{tabular}
}
\end{table}
\definecolor{mygray}{HTML}{EFEFEF}
\begin{table}[t]
\centering
\caption{\textbf{Controlled comparison of representation and injection on LIBERO-PRO.} Given the identical oracle 2D target point, 2D grounding signals are fundamentally limited regardless of injection mechanism, while 3D representation unlocks its full potential only via our direct injection.}
\label{tab:controlled_experiment}
\resizebox{0.72\linewidth}{!}{
\begin{tabular}{l ccc ccc}
\toprule
\multirow{2}{*}{\textbf{Method}} & \multicolumn{3}{c}{Task Perturbation} & \multicolumn{3}{c}{Position Perturbation} \\
\cmidrule(lr){2-4} \cmidrule(lr){5-7}
  & Object & Spatial & Avg. & Object & Spatial & Avg. \\
\midrule

GR00T-N1.6~\cite{gr00tn1_2025} & 10.0 & 52.4 & 31.2 & 29.2 & 27.0 & 28.1 \\
\midrule
+ 2D via Text Prompt & 21.6 & 63.4 & 42.5 & 42.4 & 21.0 & 31.7 \\
+ 2D via Visual Prompt & 15.8 & 64.8 & 40.3 & 43.0 & 17.8 & 30.4 \\
+ 2D via AdaLN & 26.4 & 69.0 & 47.7 & 44.0 & 31.6 & 37.8 \\
+ 3D via Text Prompt & 16.2 & 60.8 & 38.5 & 49.4 & 29.4 & 39.4 \\
+ 3D via AdaLN (Ours) & \textbf{70.4} & \textbf{84.6} & \textbf{77.5} & \textbf{79.4} & \textbf{41.0} & \textbf{60.2} \\
\bottomrule
\end{tabular}
}
\end{table}
Tab.~\ref{tab:libero_and_libero_pro} reports results of our method on LIBERO and LIBERO-PRO. Alongside our two controlled-comparison backbones, we include three external models as reference points: OpenVLA~\cite{kim2024openvla}, $\pi_{0}$~\cite{pi0}, and the reasoning-based MolmoAct~\cite{molmoact2025}. Although all baselines achieve strong in-distribution success, performance collapses under perturbation: OpenVLA and $\pi_{0}$ drop to $24.1$ and $32.2$ points under Task Perturbation, and to $0.0$ and $5.9$ points under Position Perturbation; even $\pi_{0.5}$ and GR00T-N1.6, trained on the same data as our method, only reach $37.3$ and $31.2$ points under Task Perturbation, and $47.0$ and $28.1$ points under Position Perturbation. This indicates that brittleness to spatial and task shifts is a widespread failure mode of current VLAs. Adding our method reverses this trend. On $\pi_{0.5}$, success rates rise to $\textbf{75.9}$ and $\textbf{72.2}$ points under Task and Position Perturbation; on GR00T-N1.6, success rates rise to $\textbf{77.5}$ and $\textbf{60.2}$ points. Consistent improvements across two backbones demonstrate that our method is backbone-agnostic across VLAs with diffusion-based action heads. 

We further investigate whether the gains simply stem from incorporating 3D information by comparing against FALCON~\cite{falcon}, which incorporates dense 3D features into the action head. Trained under our 69-task setting, FALCON shows limited generalization under perturbations, particularly under Position Perturbation. We hypothesize that dense 3D features make the policy more prone to overfitting: the 3D features shift under position perturbation, and the policy fails to generalize to this. In contrast, our grounding-aware 3D representations focus on task-relevant spatial information, leading to better generalization. These results suggest that the gains come not merely from depth information, but from how 3D representations are grounded for action prediction.


\subsection{Disentangling Grounding Representation and Injection}
\label{sec:grounding_rep_inj}

The results in Sec.~\ref{sec:libero_and_libero_pro} conflate two design choices: representing the grounding signal in 3D and injecting it directly into the action head. To isolate the contribution of each, we design a controlled comparison spanning these two axes. All variants receive the \emph{identical} oracle 2D target point and differ only in how this signal is represented and injected: \textbf{2D via Text Prompt} appends the 2D point coordinates to the text prompt; \textbf{2D via Visual Prompt} overlays a marker on the input image; \textbf{2D via AdaLN} encodes the tuple $(x_t, y_t,\, x_t - x_g,\, y_t - y_g)$ via an MLP and injects it through AdaLN, where $(x_t, y_t)$ is the target point and $(x_g, y_g)$ is the 2D image projection of the gripper; \textbf{3D via Text Prompt} appends the lifted 3D coordinates to the text prompt; and \textbf{3D via AdaLN (Ours)} is our method. Full implementation details are provided in App.~\ref{supp:sec:subsec:grounding_variants}. Results are reported in Tab.~\ref{tab:controlled_experiment}.

\textbf{2D grounding signals are fundamentally limited regardless of injection mechanism.} All three 2D variants yield slight gains over the baseline, and even 2D via AdaLN—which adopts the same injection mechanism as ours—still falls noticeably short. Since action prediction operates in 3D physical space, a 2D signal forces the policy to internally learn the 2D-to-3D mapping—a burden the VLA struggles to absorb regardless of how the 2D signal is conveyed. Supplying the signal directly in 3D removes this burden and is therefore the more effective choice.

\textbf{3D representation unlocks its full potential via direct injection.} We observe that 3D via Text Prompt yields no clear improvement over the three 2D variants, indicating that lifting the signal into 3D alone is not enough—the injection mechanism matters equally. When 3D coordinates are appended to the text prompt, they are tokenized and processed by a VLM backbone that has no strong prior for interpreting numerical 3D coordinates, and the geometric structure is largely lost before reaching the action head. Injecting the 3D displacement as a continuous embedding through AdaLN instead delivers the geometric signal directly to the action head, preserving its full fidelity. 

These findings show that the two design choices of our method are \emph{complementary}: 3D representation provides a signal whose geometry matches the action space, and direct injection ensures this geometry reaches the action head intact.

\subsection{Design Choices Outweigh Data Scaling}
\label{sec:tasks_scaling}
\begin{figure}[t]
\centering
\includegraphics[width=0.85\linewidth]{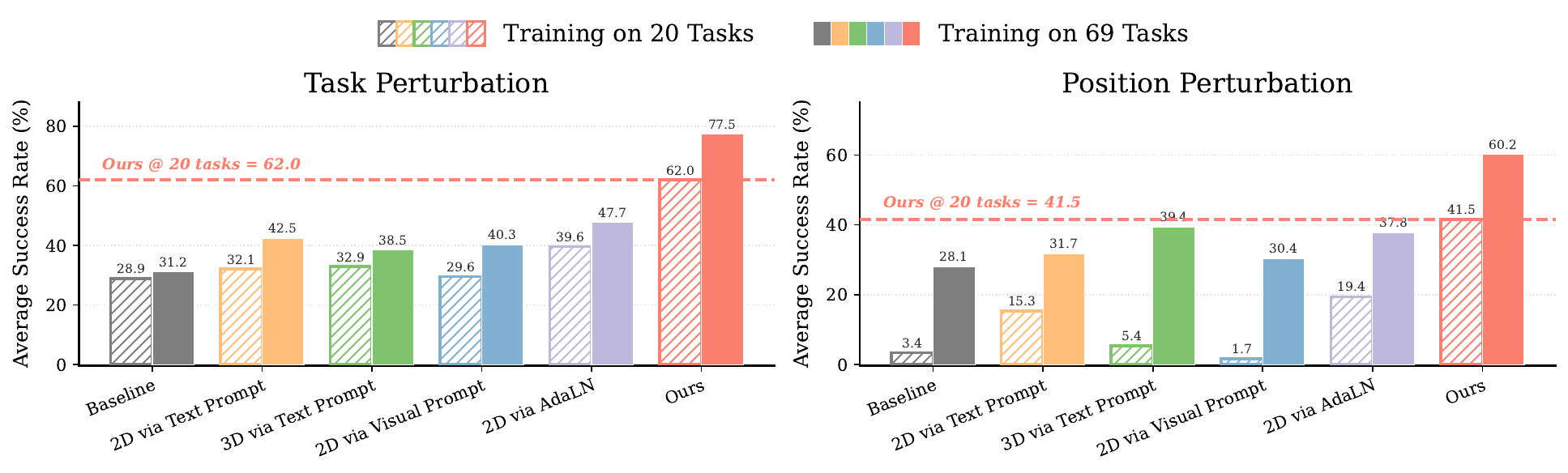}
\caption{\textbf{Comparison of all variants at 20-task and 69-task training scales.} Our method trained on only 20 tasks already surpasses every other method trained on 69 tasks, indicating a well-chosen design outweighs data scaling.}
\label{fig:tasks_scaling_exp}
\end{figure}

All previous results are obtained with the 69-task training set. In this section, we retrain all methods on the original 20-task setup (LIBERO-Object + LIBERO-Spatial). Results are shown in Fig.~\ref{fig:tasks_scaling_exp}. Expanding the training set from 20 to 69 tasks yields consistent gains across the baseline and all controlled variants, confirming that broader task diversity is a meaningful source of generalization. However, even with the larger training set, none of these methods matches the performance that our method already attains with the smaller 20-task setup. This suggests that while data diversity does contribute to generalization, a well-chosen design is a far more effective lever.

\subsection{Robustness to Noisy Depth Inputs}
\label{sec:depth_noise_exp}
\begin{wrapfigure}{r}{0.5\textwidth}
    \includegraphics[width=\linewidth]{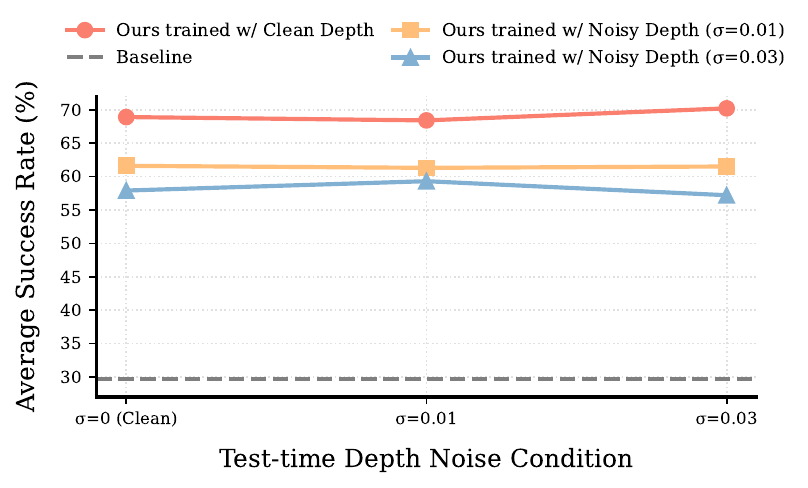}
    \caption{\textbf{Depth noise experiments.} Models trained with noisy depth show a moderate drop but still outperform the baseline, and our method is tolerant to depth noise at test time.}
    \label{fig:depth_noise_exp}
\end{wrapfigure}

We evaluate robustness of our method to depth noise by adding Gaussian noise ($\sigma = 0.01$ and $0.03$ m) to depth during training and testing. Fig.~\ref{fig:depth_noise_exp} reports the average success rate across LIBERO-Object and LIBERO-Spatial under Task and Position Perturbation.



\noindent\textbf{Effect of Training-Time Depth Noise.} Training with noisy depth causes moderate degradation compared to the clean-depth variant, as noise propagates to errors in the estimated 3D positions. Nevertheless, both noisy-training variants substantially outperform the baseline, showing that our method retains its advantages even under imperfect training conditions.

\noindent\textbf{Effect of Test-Time Depth Noise.} All three variants---regardless of whether they were trained on clean or noisy depth---maintain comparable performance across all noise levels, indicating that our method is inherently robust to depth perturbations at inference time.

\subsection{Robustness to Noisy Grounding Inputs}
\label{sec:gnd_noise_exp}
\begin{wrapfigure}{r}{0.5\textwidth}
    \includegraphics[width=\linewidth]{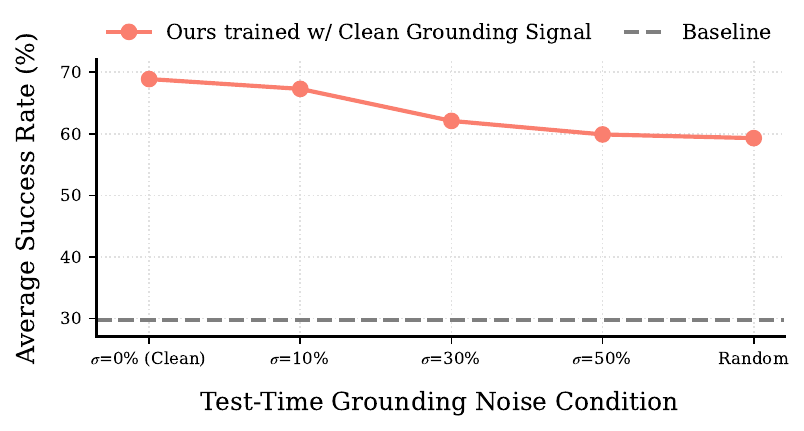}
    \caption{\textbf{Grounding noise experiments.} Although performance degrades with grounding noise, our method substantially outperforms the baseline even under severe noise.}
    \label{fig:grounding_noise_exp}
\end{wrapfigure} 
We evaluate the robustness of our method to grounding noise by perturbing the oracle 2D point at test time with Gaussian pixel noise, whose standard deviation $\sigma$ is set to a fraction of the target object's bounding-box size. We re-sample the point until it falls within the object mask and also consider an extreme setting that randomly samples 2D points from the mask. 
Fig.~\ref{fig:grounding_noise_exp} reports the average success rate across LIBERO-Object and LIBERO-Spatial under Task and Position Perturbation.
Performance gradually degrades with increasing noise, but our method consistently outperforms the baseline by a substantial margin, even under severe noise ($\sigma=50\%$) or random grounding. These results demonstrate the robustness of our method to imperfect grounding within the target object at inference time.

\subsection{Ablation Study}
\begin{wraptable}{r}{0.33\textwidth}
\centering
\captionof{table}{Spatial embedding input components ablation.}
\label{tab:ablation_inputs}
\resizebox{\linewidth}{!}{
\begin{tabular}{ >{\centering\arraybackslash}p{1.2cm} >{\centering\arraybackslash}p{1.0cm} >{\centering\arraybackslash}p{1.0cm} c }
\toprule
\multicolumn{3}{c}{\textbf{Spatial Embedding Inputs}} & \multirow{2}{*}{\textbf{Avg. Success Rate}} \\
\cmidrule(lr){1-3}
$\mathbf{p}_t$ & $\mathbf{p}_g$ & $\mathbf{d}$ & \\
\midrule
\multicolumn{3}{c}{\textcolor{gray}{GR00T-N1.6 (Baseline)}} & \textcolor{gray}{29.7} \\
\midrule
\checkmark &            &            & 34.3 \\
\checkmark & \checkmark &            & 53.7 \\
           &            & \checkmark & \textbf{68.9} \\
\checkmark & \checkmark & \checkmark & 59.7 \\
\bottomrule
\end{tabular}}

\captionof{table}{Injection mechanism ablation.}
\label{tab:ablation_injection}
\resizebox{\linewidth}{!}{
\begin{tabular}{l c}
\toprule
\textbf{Injection Method} & \textbf{Avg. Success Rate} \\
\midrule
\textcolor{gray}{GR00T-N1.6 (Baseline)} & \textcolor{gray}{29.7} \\
\midrule
Robot State    & 51.7 \\
In-Context~\cite{DiT}    & 58.5 \\
AdaLN (Ours)   & \textbf{68.9} \\
\bottomrule
\end{tabular}}
\end{wraptable}
We ablate the spatial embedding input components and its injection mechanism into the action head on top of GR00T-N1.6, reporting the average success rate across LIBERO-Object and LIBERO-Spatial under Task and Position Perturbation.

\noindent\textbf{Spatial Embedding Input Components.}
As shown in Tab.~\ref{tab:ablation_inputs}, we observe that the 3D grounding signal becomes useful when paired with the gripper's position, indicating that the spatial embedding learns the gripper-to-target geometry. Besides, we observe that $\mathbf{d}$ alone outperforms the combined $(\mathbf{p}_t, \mathbf{p}_g, \mathbf{d})$ variant. We attribute this to the absolute positions varying across scenes in LIBERO (e.g. table heights), making it harder to learn an embedding that transfers across scenes and tasks. Using only $\mathbf{d}$ sidesteps this by injecting gripper-to-target geometry as an explicit inductive bias, exposing the policy to a scene-invariant signal and yielding stronger generalizability.

\noindent\textbf{Injection Mechanism.}
We compare AdaLN-based injection against two self-attention-based alternatives. The first injects the displacement $\mathbf{d}$ through the robot state, concatenating it with the proprioceptive input and encoding it as a state token. The second adopts the In-Context conditioning strategy from DiT~\cite{DiT}, appending a spatial embedding as an extra token to the action and state tokens. As shown in Tab.~\ref{tab:ablation_injection}, In-Context conditioning outperforms the robot-state variant, but both remain inferior to AdaLN, suggesting that directly modulating normalized features provides a more effective pathway for spatial conditioning.

\section{Real-World Experiments}
\subsection{Real-World Experimental Setup}
\begin{wrapfigure}{r}{0.55\textwidth}
    \includegraphics[width=\linewidth]{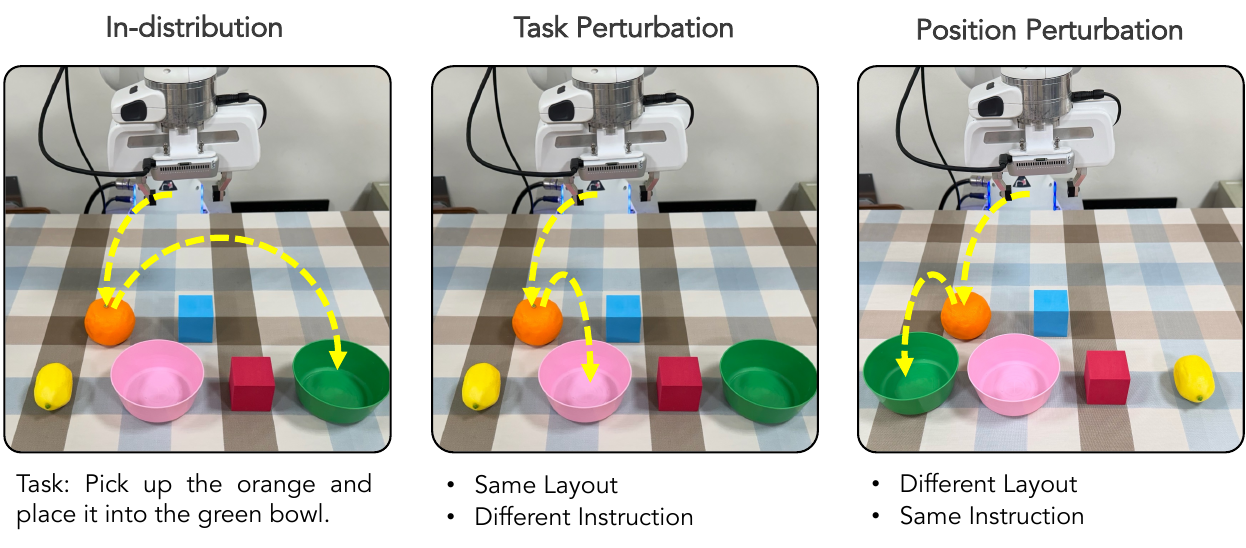}
    \caption{\textbf{Real-world experiment task setup}. Each task is evaluated under three conditions: In-Distribution, Task Perturbation, and Position Perturbation, to assess the policy's generalization across task and spatial variations in the real world.}
    \label{fig:real_world_exp_setup}
\end{wrapfigure}

\paragraph{Platform.} We use a Franka Emika Panda robot with two RealSense D435 cameras to  capture a front view (RGB-D) and a wrist-mounted view (RGB).

\noindent\textbf{Tasks and Evaluation.} We collect 8 pick-and-place tasks and 40 demonstrations per task for training. As shown in Fig.~\ref{fig:real_world_exp_setup}, each task is evaluated under three conditions: \textit{In-Distribution} uses the original instruction and layout; \textit{Task Perturbation} keeps the layout but swaps the target or placement region with another object, forming an unseen instruction–scene pairing; \textit{Position Perturbation} keeps the instruction but relocates the target or placement region. We run 10 rollouts per condition per task and report the average success rate. For more details, please refer to App.~\ref{supp:sec:subsec:real-world-task-setup}.

\noindent\textbf{Base Model and Baselines.} We apply our method to two pre-trained VLA backbones, $\pi_{0.5}$~\cite{pi05} and GR00T-N1.6~\cite{gr00tn1_2025}. We further compare against different representation and injection variants described in Sec.~\ref{sec:grounding_rep_inj}, all implemented using GR00T-N1.6 as the backbone.

\noindent\textbf{Grounding Source and Sub-Goal Switching.} We use Qwen3-VL-4B~\cite{qwen3vl} as the grounding source. We record the target object's z-coordinate at episode start and switch the active target from object to placement region once it rises by more than 1 cm. For more details, please refer to App.~\ref{supp:sec:subsec:real-world-grounding-source}.

\subsection{Real-World Experimental Results}
\label{sec:subsec:real-world-experimental-results}
\definecolor{mygray}{HTML}{EFEFEF}
\definecolor{myblue}{HTML}{EAF3FF}
\definecolor{deltagreen}{HTML}{3A7D44}
\definecolor{deltared}{HTML}{CC3333}

\begin{table}[t]
\centering
\caption{\textbf{Real-world experiment results.} Both backbones collapse under Task and Position Perturbation, while our method restores most of the performance and outperforms other grounding-augmented variants. The best performance in each column is highlighted in bold.}

\label{tab:real_exp}
\resizebox{0.75\linewidth}{!}{
\begin{tabular}{l ccc}
\toprule
\textbf{Method} & In Distribution & Task Perturbation & Position Perturbation \\
\midrule

$\pi_{0.5}$~\cite{pi05} & 85.0 & 32.5 & 27.5 \\
\rowcolor{myblue} $\pi_{0.5}$ + Ours & 91.3 & \textbf{78.8} & 61.3 \\
\rowcolor{myblue} $\Delta$ & \textcolor{deltagreen}{\footnotesize{6.3 $\uparrow$}} & \textcolor{deltagreen}{\footnotesize{46.3 $\uparrow$}} & \textcolor{deltagreen}{\footnotesize{33.8 $\uparrow$}} \\
\midrule
GR00T-N1.6~\cite{gr00tn1_2025} & 68.8 & 3.8 & 3.8 \\
\rowcolor{myblue} GR00T-N1.6 + Ours & \textbf{92.5} & 63.8 & \textbf{66.3} \\
\rowcolor{myblue} $\Delta$ & \textcolor{deltagreen}{\footnotesize{23.7 $\uparrow$}} & \textcolor{deltagreen}{\footnotesize{60.0 $\uparrow$}} & \textcolor{deltagreen}{\footnotesize{62.5 $\uparrow$}} \\
\midrule
\rowcolor{mygray} \multicolumn{4}{l}{\textit{Controlled comparison of representation and injection}} \\
\midrule
GR00T-N1.6 + 2D via Text Prompt & 87.5 & 5.0 & 7.5 \\
GR00T-N1.6 + 2D via Visual Prompt & 88.8 & 6.3 & 5.0 \\
GR00T-N1.6 + 2D via AdaLN & 87.5 & 18.8 & 16.3 \\
GR00T-N1.6 + 3D via Text Prompt & 86.3 & 5.0 & 5.0 \\
\bottomrule
\end{tabular}}
\end{table}
Tab.~\ref{tab:real_exp} reports the results on 8 real-world tasks. 
Our method improves both backbones under both perturbations. On $\pi_{0.5}$, success rates rise from $32.5$ and $27.5$ to $78.8$ and $61.3$ points under Task and Position Perturbation; on GR00T-N1.6, from $3.8$ and $3.8$ to $63.8$ and $66.3$. Notably, our real-world setup obtains 2D grounding from an off-the-shelf VLM and depth from a consumer RGB-D sensor---both substantially noisier than the oracle signals in simulation---yet the gains persist, confirming the practical applicability of our method.

The controlled comparison on GR00T-N1.6 reproduces the trend of Sec.~\ref{sec:grounding_rep_inj}. In distribution, every grounding-augmented variant improves over the baseline; the designs separate only under perturbation. There, both 2D prompting variants and 3D via Text Prompt collapse to near-baseline success, and even the strongest alternative, 2D via AdaLN, reaches only $18.8$ and $16.3$, far behind ours. Consistent with our analysis in Sec.~\ref{sec:grounding_rep_inj}, 2D via AdaLN still requires the policy to learn the 2D-to-3D mapping, whereas our signal already lies in 3D space, leading to better generalization. As in simulation, only the combination of a 3D representation and direct injection into the action head enables transfer from in-distribution to out-of-distribution conditions.

\section{Discussion and Limitation}
In this work, we revisit how spatial grounding should be injected into VLAs and identify a simple but overlooked finding: given an adequate 2D grounding signal, lifting it into 3D and injecting it directly into the action head is what unlocks both spatial and instance-level task generalization. Built on this insight, our lightweight, model-agnostic method delivers substantial gains on both GR00T-N1.6 and $\pi_{0.5}$ on LIBERO-PRO benchmark, and remains effective in real-world settings.

\textbf{Limitations.} Our method rests on three assumptions that limit its scope. First, we assume each sub-goal is associated with a target object or region, leaving object-agnostic motions outside the formulation; lifting 2D trajectories into 3D and injecting them through a similar mechanism is a possible extension. Second, we assume an adequate 2D grounding signal that identifies the target, so the task generalization we report is instance-level within pick-and-place: faithfully acting on this target under unseen instruction–scene pairings, not generalizing to new skills. And we use relatively simple grounding sources to isolate the effect of representation and injection; robust grounding for more complex scenes or instructions remains an open problem orthogonal to our contribution. Third, lifting 2D grounding into 3D requires an RGB-D camera with calibrated intrinsics and extrinsics; relaxing this, e.g., via monocular depth estimation, is left to future work.


\clearpage
\acknowledgments{We thank the National Science and Technology Council (NSTC), Taiwan, for supporting this research under grant numbers NSTC 113-2221-E-007 -105 -MY3, NSTC 114-2634-F-002-004, and NSTC 115-2640-E-007-015. We further acknowledge the National Center for High-Performance Computing (NCHC), Taiwan, for access to its computing and storage facilities.}


\bibliography{main}  
\clearpage
\appendix
\section*{Appendix}
\section{Table of Contents}
\begin{enumerate}
    \item Sec.~\ref{supp:sec:implementation_details} provides \textbf{Implementation Details} for lifting a 2D point into 3D and encoding the spatial embedding, injecting it into the VLA, and each variant evaluated in Sec.~\ref{sec:grounding_rep_inj} of the main paper.
    \item Sec.~\ref{supp:sec:simulation_setup_details} provides \textbf{Simulation Experimental Details}, including the training tasks, how we obtain 2D grounding in simulation, the model training configurations, and the full experimental results.
    \item Sec.~\ref{supp:sec:real_world_setup_details} provides \textbf{Real-World Experimental Details}, including the task and evaluation setup, how we obtain 2D grounding in real world, the model training configurations, and the full experimental results.
    \item Sec.~\ref{supp:sec:qualitative_results} provides \textbf{Qualitative Results and Real-World Failure Modes}.
    
\end{enumerate}

\section{Implementation Details}~\label{supp:sec:implementation_details}
In this section, we provide implementation details for: (1) lifting a 2D point into 3D and encoding the spatial embedding (Sec.~\ref{supp:sec:subsec:lifting}); (2) injecting the spatial embedding into the VLA (Sec.~\ref{supp:sec:subsec:app_inject}); and (3) each variant evaluated in Sec.~\ref{sec:grounding_rep_inj} of the main paper (Sec.~\ref{supp:sec:subsec:grounding_variants}).

\subsection{From 2D Target Point to 3D Spatial Embedding}
\label{supp:sec:subsec:lifting}
This subsection details the lifting procedure that converts a 2D target point on the third-person view image into a 3D position expressed in the robot base frame, and the subsequent encoding into a spatial embedding.

\subsubsection{Notation}
Let $(u, v) \in \mathbb{R}^2$ denote the 2D target pixel produced by the grounding source on the third-person view image, and $D \in \mathbb{R}^{H \times W}$ the aligned depth image of the same view. We denote the camera intrinsic matrix as $K \in \mathbb{R}^{3 \times 3}$, with focal lengths $(f_x, f_y)$ and principal point $(c_x, c_y)$, and the camera extrinsic as $T^{\text{base}}_{\text{cam}} = (R, \mathbf{t}) \in \mathrm{SE}(3)$, which maps a point from the camera frame to the robot base frame. The gripper position $\mathbf{p}_g \in \mathbb{R}^3$ is obtained directly from the proprioceptive robot state via forward kinematics, and is already expressed in the robot base frame.

\subsubsection{Lifting 2D to 3D}
We first read the metric depth at the target pixel, $z = D(u, v)$, and back-project the pixel to the camera frame using the pinhole model:
\begin{equation}
    \mathbf{p}^{\text{cam}}_t \;=\; z \cdot K^{-1} \begin{bmatrix} u \\ v \\ 1 \end{bmatrix}
    \;=\;
    \begin{bmatrix}
        (u - c_x)\, z / f_x \\
        (v - c_y)\, z / f_y \\
        z
    \end{bmatrix}.
\end{equation}
Applying the camera extrinsic then yields the target position in the robot base frame:
\begin{equation}
\mathbf{p}_t \;=\; R\, \mathbf{p}^{\text{cam}}_t + \mathbf{t}.
\end{equation}
\subsubsection{Compute Displacement and Spatial Embedding}
Since $\mathbf{p}_t$ and $\mathbf{p}_g$ now share the same coordinate frame, their difference is geometrically well-defined. We compute the 3D relative displacement:
\begin{equation}
\mathbf{d} = \mathbf{p}_t - \mathbf{p}_g,
\end{equation}
and encode it through a two-layer MLP with ReLU activation to obtain the spatial embedding:
\begin{equation}
    \mathbf{z}_{\text{spatial}} = \operatorname{MLP}(\mathbf{d}) \in  \mathbb{R}^{d_h},
    \label{eq:mlp}
\end{equation}
where $d_h$ is the hidden size of the action head. For both GR00T-N1.6~\cite{gr00tn1_2025} and $\pi_{0.5}$~\cite{pi05}, the MLP's hidden and output dimensions are both set to 1024.

\subsection{Inject Spatial Embedding into the VLA}
\label{supp:sec:subsec:app_inject}

The VLA backbones we build upon adopt a DiT-based action head whose blocks are modulated through an adaptive layer normalization (AdaLN) conditioning mechanism~\cite{DiT, perez2018film}. In its original formulation, AdaLN conditions each DiT block solely on the diffusion timestep embedding $\mathbf{z}_{\text{time}} \in \mathbb{R}^{d_h}$, where a per-block linear projection regresses scale and shift parameters that modulate the normalized features:
\begin{equation}
    \gamma_{\text{orig}},\ \beta_{\text{orig}}
    = \operatorname{Linear}(\mathbf{z}_{\text{time}}).
    \label{eq:adaln_orig}
\end{equation}
To inject our spatial signal, we extend this mechanism by fusing $\mathbf{z}_{\text{spatial}}$ with $\mathbf{z}_{\text{time}}$ via element-wise addition before regressing $\gamma$ and $\beta$:
\begin{equation}
    \gamma,\ \beta
    = \operatorname{Linear}(\mathbf{z}_{\text{time}}
      + \mathbf{z}_{\text{spatial}}).
    \label{eq:adaln_ours}
\end{equation}
The resulting parameters then modulate the normalized features in each DiT block of the action head, allowing the policy to incorporate the task-relevant 3D geometric information carried by $\mathbf{z}_{\text{spatial}}$ when predicting actions. Since the existing AdaLN linear projection is reused as-is, no additional parameters are introduced inside the action head.

\noindent\textbf{Preserving pretrained AdaLN behavior at initialization.}
A naive instantiation of Eq.~\ref{eq:adaln_ours} perturbs the pretrained AdaLN behavior from the very first training step: as soon as $\mathbf{z}_{\text{spatial}}$ is added to $\mathbf{z}_{\text{time}}$, the input to the AdaLN linear projection shifts away from the distribution it was pretrained on, which can destabilize the pretrained timestep conditioning that the action head relies on. To avoid this, we initialize the weight and bias of the \emph{second} layer of the spatial MLP (Eq.~\ref{eq:mlp}) to zero, while the first layer retains its standard initialization. With this choice, $\mathbf{z}_{\text{spatial}} = \mathbf{0}$ at initialization, so $\mathbf{z}_{\text{time}} + \mathbf{z}_{\text{spatial}} = \mathbf{z}_{\text{time}}$ and Eq.~\ref{eq:adaln_ours} exactly recovers the pretrained $\gamma_{\text{orig}}, \beta_{\text{orig}}$ of Eq.~\ref{eq:adaln_orig} at training step~0. Training therefore begins from the exact pretrained behavior of the action head, and the spatial signal is incorporated only gradually as the second-layer parameters move away from zero through gradient updates.  Empirically, this zero-initialization strategy preserves the pretrained timestep conditioning at the start of fine-tuning and provides a stable starting point from which the action head learns to exploit the spatial signal.

\subsection{Different Grounding Representation and Injection Variants Implementation Details}
\label{supp:sec:subsec:grounding_variants}
\newtcolorbox{promptbox}{
  colback=gray!5,
  colframe=gray!50,
  boxrule=0.5pt,
  arc=1mm,
  left=2mm, right=2mm, top=1mm, bottom=1mm,
  fontupper=\small\ttfamily,
}
\begin{wrapfigure}{r}{0.5\textwidth}
    \includegraphics[width=\linewidth]{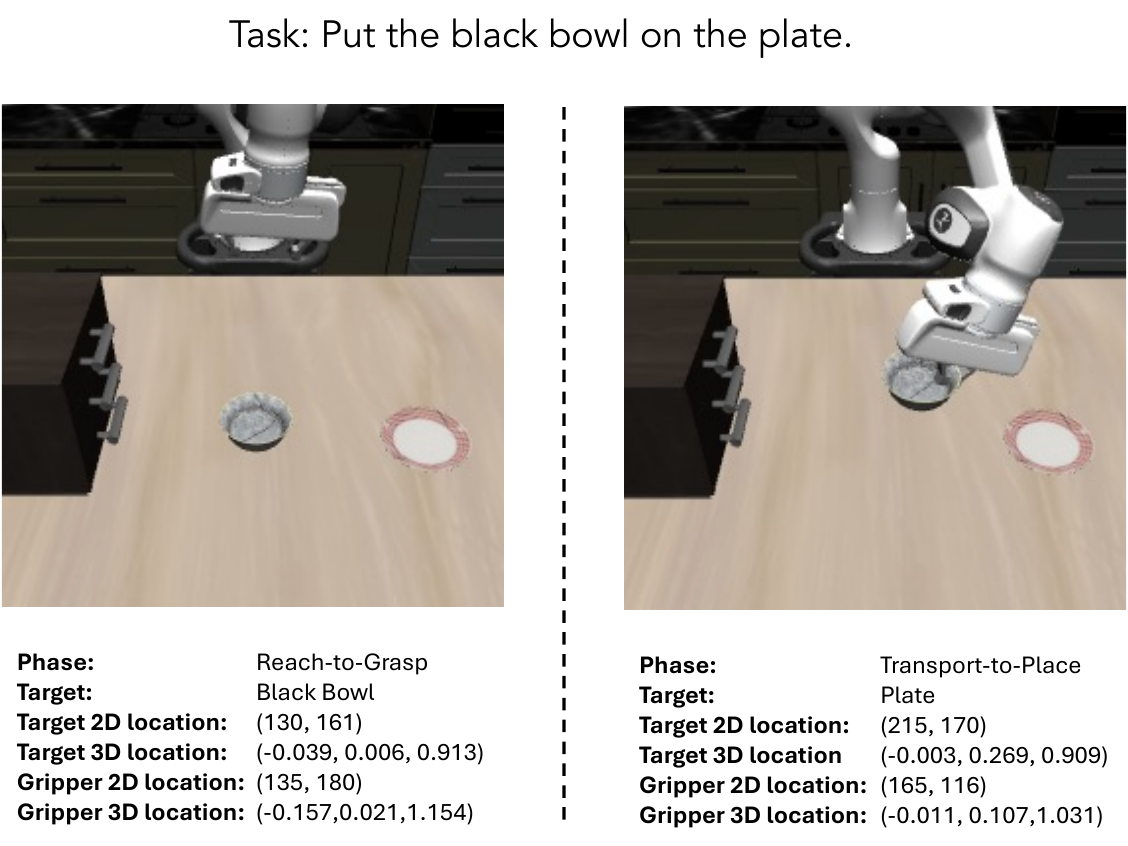}
    \caption{Example pick-and-place demo showing the target object's and gripper's 2D and 3D coordinates across the two different phases.}
    \label{fig:pick_and_place_frame}
\end{wrapfigure}
In this subsection, we provide the implementation details of each variant evaluated in Sec.~\ref{sec:grounding_rep_inj}. In Fig.~\ref{fig:pick_and_place_frame}, we present an example demo showing the target’s 2D and 3D coordinates across two frames corresponding to reach-to-grasp and transport-to-place phases in a pick-and-place task. In the following paragraphs, we use these two frames from the demo to illustrate how each variant is implemented.

\paragraph{2D via Text Prompt Variant.}
For the 2D via Text Prompt variant, we append the 2D pixel coordinates to the language prompt. We experiment with two normalization schemes, scaling the coordinates to either $[0, 1]$ or $[0, 1000]$, and find that the former performs better. Therefore, all subsequent experiments adopt the $[0, 1]$ normalization, where coordinates are divided by the image width and height. For the reach-to-grasp phase, we use the following prompt:

\begin{promptbox}
Put the black bowl on the plate. Pick up the object at [0.51, 0.63].
\end{promptbox}

For the transport-to-place phase, we use the following prompt:
\begin{promptbox}
Put the black bowl on the plate. Place the object around [0.84, 0.66].
\end{promptbox}

\paragraph{2D via Visual Prompt Variant.}
\begin{wrapfigure}{r}{0.5\textwidth}
    \includegraphics[width=\linewidth]{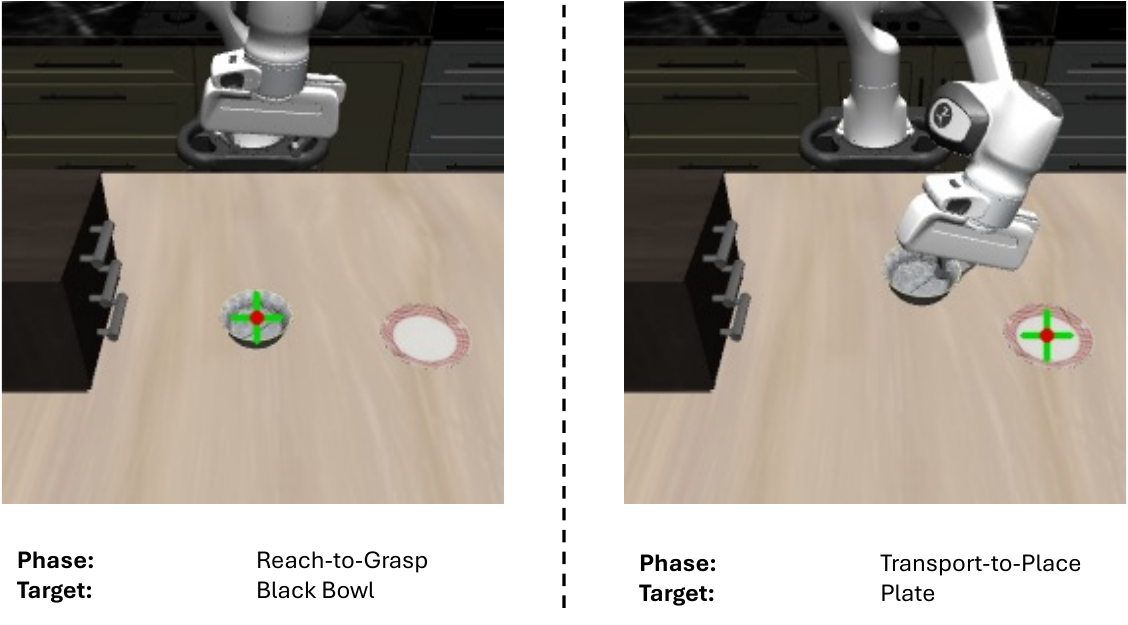}
    \caption{Illustration of the Visual Prompt variant.}
    \label{fig:visual_prompt}
\end{wrapfigure}
For the 2D via Visual Prompt variant, we overlay a point marker on the target location in the input image. As shown in Fig.~\ref{fig:visual_prompt}, we use a cross-shaped marker, following the style of prior work~\cite{wang2026vp}, to indicate the target location. In the text prompt, we additionally append an instruction specifying that the model should pick up or place the object at the location indicated by the marker. For the reach-to-grasp phase, we use the following prompt:

\begin{promptbox}
Put the black bowl on the plate. Pick up the object indicated by the cross marker.
\end{promptbox}

For the transport-to-place phase, we use the following prompt:
\begin{promptbox}
Put the black bowl on the plate. Place the object around the location indicated by the cross marker.
\end{promptbox}

\paragraph{2D via AdaLN Variant.}
For the 2D via AdaLN variant, we first project the gripper position into the third-person camera frame to obtain the gripper 2D position $(x_g, y_g)$. Given the target 2D position $(x_t, y_t)$, we encode the tuple $(x_t, y_t, x_t - x_g, y_t - y_g)$ with an MLP to produce an embedding, which is then injected through AdaLN. This variant is similar to our method, except that the grounding information is represented in 2D rather than 3D.

\paragraph{3D via Text Prompt Variant.} 
For the 3D via Text Prompt variant, we append the target 3D location in the robot base frame to the language prompt. For the reach-to-grasp phase, we use the following prompt:

\begin{promptbox}
Put the black bowl on the plate. Pick up the object at [-0.039, 0.006, 0.913].
\end{promptbox}

For the transport-to-place phase, we use the following prompt:
\begin{promptbox}
Put the black bowl on the plate. Place the object around [-0.003, 0.269, 0.909].
\end{promptbox}
\section{Simulation Experimental Details}~\label{supp:sec:simulation_setup_details}
In this section, we provide additional details of our simulation experiments, including: (1) the training tasks in our setup (Sec.~\ref{supp:sec:subsec:training_tasks}), (2) how we obtain 2D grounding information in simulation (Sec.~\ref{supp:sec:subsec:grounding-source}), (3) the per-model training configurations (Sec.~\ref{supp:sec:subsec:sim-training-details}), and (4) the full experimental results (Sec.~\ref{supp:sec:subsec:sim-experimental-results-details}).

\subsection{Training Data Details}
\label{supp:sec:subsec:training_tasks}
Our default training setup augments LIBERO-Object and LIBERO-Spatial with 49 additional pick-and-place tasks from LIBERO-90, yielding 69 tasks in total. 
In Tab.~\ref{tab:libero90_tasks}, we list the 49 additional pick-and-place tasks.
Our selection criterion is that a mask for the target object or region must be obtainable from the observation. For abstract regions without an explicit mask (e.g., the left side of a mug), we instead ensure that the 2D projection of the region's center is not occluded. For example, tasks with target region such as the second shelf of a cabinet are excluded from our selection, since their center, when projected to 2D, falls on the surface of the top shelf and is therefore occluded.
\begin{table}[t]
\centering
\footnotesize
\caption{The 49 tasks from LIBERO-90 used in the 69-task setup.}
\label{tab:libero90_tasks}
\begin{tabular}{ll}
\toprule
Scene & Task Instruction \\
\midrule
\texttt{KITCHEN\_SCENE1}      & put the black bowl on the plate \\
\texttt{KITCHEN\_SCENE1}      & put the black bowl on top of the cabinet \\
\texttt{KITCHEN\_SCENE2}      & put the black bowl at the back on the plate \\
\texttt{KITCHEN\_SCENE2}      & put the black bowl at the front on the plate \\
\texttt{KITCHEN\_SCENE2}      & put the middle black bowl on the plate \\
\texttt{KITCHEN\_SCENE2}      & put the middle black bowl on top of the cabinet \\
\texttt{KITCHEN\_SCENE2}      & stack the black bowl at the front on the black bowl in the middle \\
\texttt{KITCHEN\_SCENE2}      & stack the middle black bowl on the back black bowl \\
\texttt{KITCHEN\_SCENE3}      & put the frying pan on the stove \\
\texttt{KITCHEN\_SCENE3}      & put the moka pot on the stove \\
\texttt{KITCHEN\_SCENE4}      & put the black bowl in the bottom drawer of the cabinet \\
\texttt{KITCHEN\_SCENE4}      & put the black bowl on top of the cabinet \\
\texttt{KITCHEN\_SCENE4}      & put the wine bottle in the bottom drawer of the cabinet \\
\texttt{KITCHEN\_SCENE5}      & put the black bowl in the top drawer of the cabinet \\
\texttt{KITCHEN\_SCENE5}      & put the black bowl on the plate \\
\texttt{KITCHEN\_SCENE5}      & put the black bowl on top of the cabinet \\
\texttt{KITCHEN\_SCENE5}      & put the ketchup in the top drawer of the cabinet \\
\texttt{KITCHEN\_SCENE6}      & put the yellow and white mug to the front of the white mug \\
\texttt{KITCHEN\_SCENE7}      & put the white bowl on the plate \\
\texttt{KITCHEN\_SCENE7}      & put the white bowl to the right of the plate \\
\texttt{KITCHEN\_SCENE8}      & put the right moka pot on the stove \\
\texttt{KITCHEN\_SCENE9}      & put the frying pan on top of the cabinet \\
\texttt{KITCHEN\_SCENE9}      & put the white bowl on top of the cabinet \\
\texttt{KITCHEN\_SCENE10}     & put the black bowl in the top drawer of the cabinet \\
\texttt{LIVING\_ROOM\_SCENE1} & pick up the alphabet soup and put it in the basket \\
\texttt{LIVING\_ROOM\_SCENE1} & pick up the cream cheese box and put it in the basket \\
\texttt{LIVING\_ROOM\_SCENE1} & pick up the ketchup and put it in the basket \\
\texttt{LIVING\_ROOM\_SCENE1} & pick up the tomato sauce and put it in the basket \\
\texttt{LIVING\_ROOM\_SCENE2} & pick up the butter and put it in the basket \\
\texttt{LIVING\_ROOM\_SCENE2} & pick up the milk and put it in the basket \\
\texttt{LIVING\_ROOM\_SCENE2} & pick up the orange juice and put it in the basket \\
\texttt{LIVING\_ROOM\_SCENE2} & pick up the tomato sauce and put it in the basket \\
\texttt{LIVING\_ROOM\_SCENE3} & pick up the alphabet soup and put it in the tray \\
\texttt{LIVING\_ROOM\_SCENE3} & pick up the butter and put it in the tray \\
\texttt{LIVING\_ROOM\_SCENE3} & pick up the cream cheese and put it in the tray \\
\texttt{LIVING\_ROOM\_SCENE3} & pick up the ketchup and put it in the tray \\
\texttt{LIVING\_ROOM\_SCENE3} & pick up the tomato sauce and put it in the tray \\
\texttt{LIVING\_ROOM\_SCENE4} & pick up the black bowl on the left and put it in the tray \\
\texttt{LIVING\_ROOM\_SCENE4} & pick up the salad dressing and put it in the tray \\
\texttt{LIVING\_ROOM\_SCENE4} & pick up the chocolate pudding and put it in the tray \\
\texttt{LIVING\_ROOM\_SCENE5} & put the red mug on the left plate \\
\texttt{LIVING\_ROOM\_SCENE5} & put the red mug on the right plate \\
\texttt{LIVING\_ROOM\_SCENE5} & put the white mug on the left plate \\
\texttt{LIVING\_ROOM\_SCENE5} & put the yellow and white mug on the right plate \\
\texttt{LIVING\_ROOM\_SCENE6} & put the chocolate pudding to the left of the plate \\
\texttt{LIVING\_ROOM\_SCENE6} & put the chocolate pudding to the right of the plate \\
\texttt{LIVING\_ROOM\_SCENE6} & put the red mug on the plate \\
\texttt{LIVING\_ROOM\_SCENE6} & put the white mug on the plate \\
\texttt{STUDY\_SCENE4}        & pick up the book on the left and place it on top of the shelf \\
\bottomrule
\end{tabular}
\end{table}

\subsection{Grounding Source Details}
\label{supp:sec:subsec:grounding-source}

We obtain 2D oracle target points from simulator-provided segmentation and object-center information, then lift them into 3D following Sec.~\ref{supp:sec:subsec:lifting}. A key invariant underlies our sampling: every 3D target point must be obtainable from a camera-visible 2D pixel via depth-based lifting, never taken directly as a 3D centroid from the simulator. This guarantees that our pipeline genuinely starts from a 2D grounding signal—the same signal an off-the-shelf VLM or detector could provide—rather than relying on privileged 3D information unavailable at deployment.

To respect this invariant, we adopt two sampling procedures depending on whether the sub-goal target admits an explicit segmentation mask. For object-typed targets, such as the target object or placement regions like \textit{plate} or \textit{tray}, the mask is directly available, and we apply a distance transform~\cite{distancetransform} to select the innermost pixel, i.e., the one with maximum distance to the mask boundary. For abstract placement regions that lack an explicit mask, such as \textit{to the left of the plate}, we instead start from the region's 3D centroid, project it onto the image to obtain a 2D pixel, and then re-lift this pixel into 3D via the same depth-based procedure. The resulting target point is the camera-observed surface point along the ray through the projected centroid, ensuring that even abstract regions are grounded through a visible 2D pixel rather than a free-space coordinate.

\subsection{Training Details}
\label{supp:sec:subsec:sim-training-details}
We detail the fine-tuning setups of GR00T-N1.6~\cite{gr00tn1_2025} and $\pi_{0.5}$~\cite{pi05} in simulation experiments below.

\noindent\textbf{Training Details for GR00T-N1.6.}
We fine-tune GR00T-N1.6 in simulation following the official training recipe on LIBERO dataset, with the only modification being the global batch size. The model predicts an action chunk of $16$ future steps per forward pass. We use a global batch size of $384$ and train for $20{,}000$ steps using AdamW with a peak learning rate of $1{\times}10^{-4}$, a cosine decay schedule with a warmup ratio of $0.05$ ($\sim\!1{,}000$ steps), weight decay of $1{\times}10^{-5}$, and gradient clipping at a max norm of $1.0$. For visual augmentation, input images are resized so that the shortest edge is $256$ px, randomly cropped to a square patch covering $0.95$ of the shortest edge, and then resized back to $256$ px. We further apply ColorJitter-style perturbations with brightness $0.3$, contrast $0.4$, saturation $0.5$, and hue $0.08$, while random rotation is disabled. All other hyperparameters follow the official configuration~\cite{gr00tn1_2025}.

\noindent\textbf{Training Details for $\pi_{0.5}$.}
We fine-tune $\pi_{0.5}$ in simulation following the official training recipe on LIBERO dataset. The model predicts an action chunk of $10$ future steps per forward pass. We use a global batch size of $256$ and train for $30{,}000$ steps using AdamW with a peak learning rate of $5{\times}10^{-5}$, a linear warmup schedule over the first $10{,}000$ steps and held constant thereafter, weight decay of $1{\times}10^{-10}$, and gradient clipping at a max norm of $1.0$. For visual augmentation, input images are resized (with padding) to $224{\times}224$ px; on the third-person view we apply a random crop covering $0.95$ of each spatial dimension, a bilinear resize back to $224{\times}224$ px, and a random rotation of $\pm 5^{\circ}$, while the wrist view receives no geometric augmentation. We further apply ColorJitter-style perturbations with brightness $0.3$, contrast $0.4$, and saturation $0.5$, while hue jitter is disabled. All other hyperparameters follow the official configuration~\cite{pi05}.

\subsection{Full Experimental Results}
\label{supp:sec:subsec:sim-experimental-results-details}
\definecolor{mygray}{HTML}{EFEFEF}
\begin{table}[t]
\centering
\caption{Controlled comparison of representation and injection on LIBERO-PRO under 20 training-task setup.}
\label{tab:controlled_experiment_20_tasks}
\resizebox{0.75\linewidth}{!}{
\begin{tabular}{l ccc ccc}
\toprule
\multirow{2}{*}{\textbf{Method}} & \multicolumn{3}{c}{Task Perturbation} & \multicolumn{3}{c}{Position Perturbation} \\
\cmidrule(lr){2-4} \cmidrule(lr){5-7}
  & Object & Spatial & Avg. & Object & Spatial & Avg. \\
\midrule

GR00T-N1.6~\cite{gr00tn1_2025} & 9.0 & 48.8 & 28.9 & 4.8 & 2.0 & 3.4 \\
\midrule
+ 2D via Text Prompt & 10.0 & 54.2 & 32.1 & 24.6 & 6.0 & 15.3 \\
+ 2D via Visual Prompt & 10.0 & 49.2 & 29.6 & 1.2 & 2.2 & 1.7 \\
+ 2D via AdaLN & 11.2 & 68.0 & 39.6 & 35.4 & 3.4 & 19.4 \\
+ 3D via Text Prompt & 10.0 & 55.8 & 32.9 & 7.4 & 3.4 & 5.4 \\
+ 3D via AdaLN (Ours) & \textbf{52.4} & \textbf{71.6} & \textbf{62.0} & \textbf{67.8} & \textbf{15.2} & \textbf{41.5} \\
\bottomrule
\end{tabular}
}
\end{table}
\definecolor{mygray}{HTML}{EFEFEF}
\begin{table}[t]
\centering
\caption{Depth noise experiment on LIBERO-PRO under different training/testing depth noise condition.}
\label{tab:depth_noise_experiment_table}
\resizebox{0.75\linewidth}{!}{
\begin{tabular}{l cc cc c}
\toprule
\multirow{2}{*}{\textbf{(Training, Testing)}} & \multicolumn{2}{c}{Task Perturbation} & \multicolumn{2}{c}{Position Perturbation} & \multirow{2}{*}{\textbf{Average}} \\
\cmidrule(lr){2-3} \cmidrule(lr){4-5}
  & Object & Spatial & Object & Spatial & \\
\midrule
\textcolor{gray}{GR00T-N1.6 (Baseline)} & \textcolor{gray}{10.0} & \textcolor{gray}{52.4} & \textcolor{gray}{29.2} & \textcolor{gray}{27.0} & \textcolor{gray}{29.7} \\
\midrule
(Clean depth, Clean depth)        & 70.4 & 84.6 & 79.4 & 41.0 & 68.9 \\
(Clean depth, $\sigma=0.01$)      & 70.2 & 83.4 & 78.0 & 41.8 & 68.4 \\
(Clean depth, $\sigma=0.03$)      & 72.0 & 88.2 & 77.4 & 43.0 & 70.2 \\
\midrule
($\sigma=0.01$, Clean depth)      & 55.8 & 84.0 & 61.6 & 44.8 & 61.6 \\
($\sigma=0.01$, $\sigma=0.01$)    & 53.6 & 81.4 & 61.8 & 48.2 & 61.3 \\
($\sigma=0.01$, $\sigma=0.03$)    & 54.6 & 83.6 & 62.4 & 45.2 & 61.5 \\
\midrule
($\sigma=0.03$, Clean depth)      & 55.2 & 74.6 & 61.0 & 40.8 & 57.9 \\
($\sigma=0.03$, $\sigma=0.01$)    & 56.4 & 74.4 & 62.4 & 44.0 & 59.3 \\
($\sigma=0.03$, $\sigma=0.03$)    & 54.6 & 72.4 & 59.8 & 41.8 & 57.2 \\

\bottomrule
\end{tabular}
}
\end{table}
\definecolor{mygray}{HTML}{EFEFEF}
\begin{table}[t]
\centering
\caption{Grounding noise experiment on LIBERO-PRO under different grounding noise condition.}
\label{tab:grounding_noise_experiment_table}
\resizebox{0.75\linewidth}{!}{
\begin{tabular}{l cc cc c}
\toprule
\multirow{2}{*}{\textbf{Noise Condition}} & \multicolumn{2}{c}{Task Perturbation} & \multicolumn{2}{c}{Position Perturbation} & \multirow{2}{*}{\textbf{Average}} \\
\cmidrule(lr){2-3} \cmidrule(lr){4-5}
 & Object & Spatial & Object & Spatial & \\
\midrule
\textcolor{gray}{GR00T-N1.6 (Baseline)} & \textcolor{gray}{10.0} & \textcolor{gray}{52.4} & \textcolor{gray}{29.2} & \textcolor{gray}{27.0} & \textcolor{gray}{29.7} \\
\midrule
$\sigma=0$ & 70.4 & 84.6 & 79.4 & 41.0 & 68.9 \\ 
$\sigma=10\%$ & 67.4 & 81.6 & 78.6 & 41.4 & 67.3 \\
$\sigma=30\%$ & 62.0 & 80.4 & 67.4 & 38.4 & 62.1 \\
$\sigma=50\%$ & 59.6 & 78.4 & 63.8 & 37.6 & 59.9 \\
Random sampling & 57.4 & 76.2 & 66.2 & 37.2 & 59.3 \\
\bottomrule
\end{tabular}
}
\end{table}
We report numerical performance values for all evaluations in Sec.~\ref{sec:tasks_scaling}, Sec.~\ref{sec:depth_noise_exp} and Sec.~\ref{sec:gnd_noise_exp}. Tab.~\ref{tab:controlled_experiment_20_tasks} presents the numerical results of the controlled comparison experiment on LIBERO-PRO under the 20-task setup (LIBERO-Object + LIBERO-Spatial), Tab.~\ref{tab:depth_noise_experiment_table} presents the numerical results of the depth noise experiment, and Tab.~\ref{tab:grounding_noise_experiment_table} presents the numerical results of the grounding noise experiment.
\section{Real-World Experimental Details}~\label{supp:sec:real_world_setup_details}
In this section, we provide additional details of our real-world experiments, including: (1) the task setup and evaluation details (Sec.~\ref{supp:sec:subsec:real-world-task-setup}), (2) how we obtain 2D grounding information in real-world (Sec.~\ref{supp:sec:subsec:real-world-grounding-source}), (3) the model training configurations (Sec.~\ref{supp:sec:subsec:real-world-training-details}), and (4) the full experimental results (Sec.~\ref{supp:sec:subsec:real-world-experimental-results-details}).

\subsection{Task Setup and Evaluation Details}
\label{supp:sec:subsec:real-world-task-setup}
\begin{figure}[t]
\centering
\includegraphics[width=\linewidth]{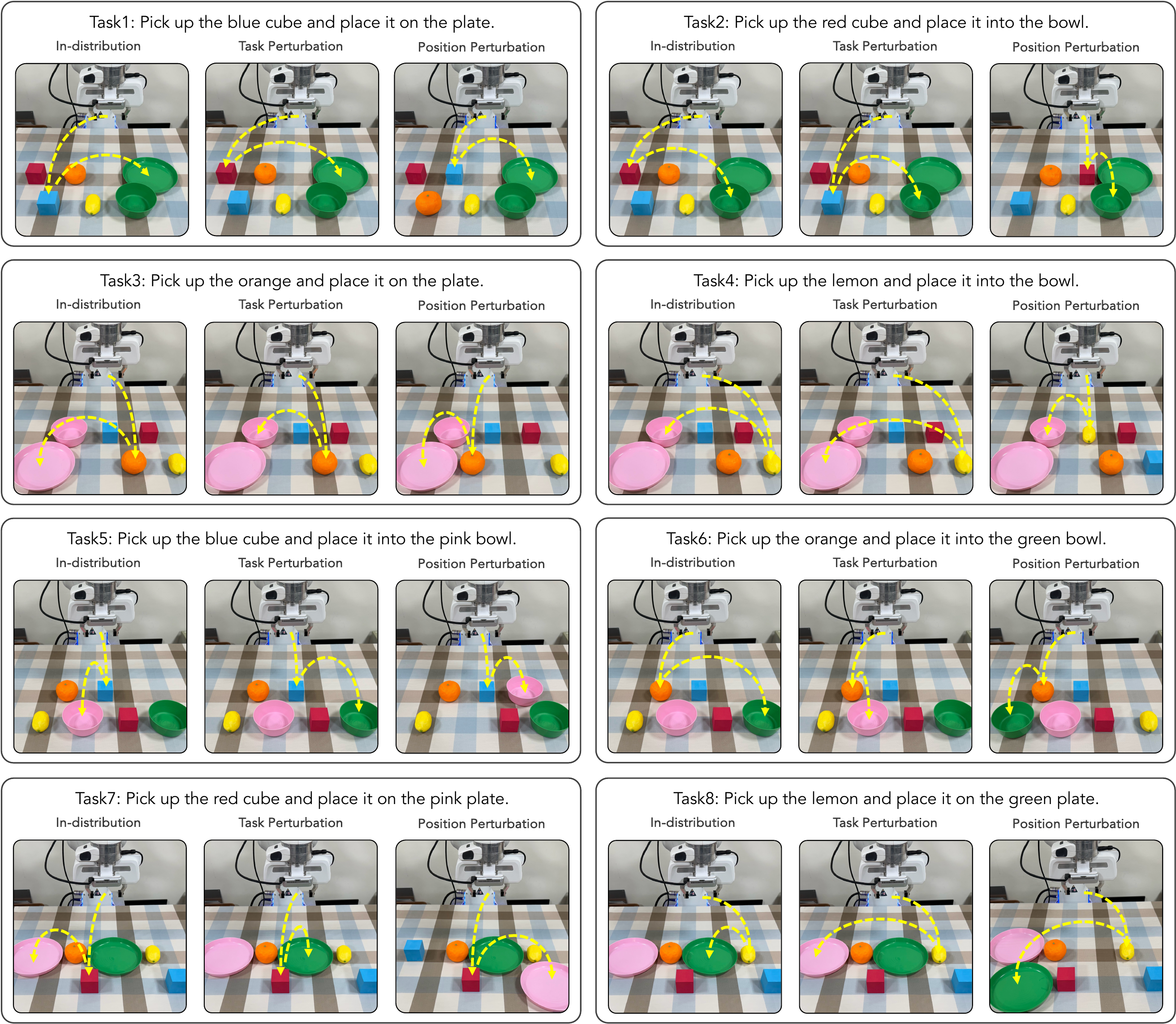}
\caption{Real-world experiment task setup.}
\label{fig:real_world_experiment_setup}
\end{figure}
We collect 8 pick-and-place tasks spanning 4 distinct object layouts, with 2 tasks sharing each layout and 40 demonstrations per task for training. The full set of tasks, together with their in-distribution (training) configurations and their test-time perturbation setups, is illustrated in Fig.~\ref{fig:real_world_experiment_setup}. Because every two tasks share a common in-distribution layout, the two tasks differ only in the instructed target object and placement region rather than in the overall scene arrangement. When collecting demonstrations, we introduce moderate positional variation across episodes while preserving the overall layout.

Each task is evaluated under three conditions. \emph{In-Distribution} uses the original training instruction and layout. \emph{Task Perturbation} keeps the layout unchanged but swaps the target object or placement region for another item, producing an instruction–scene pairing that is absent from the training set. \emph{Position Perturbation} preserves the instruction but relocates the target object or placement region. For both perturbations, the target objects and placement regions involved are themselves objects and locations that the policy has already picked from or placed into in \emph{other} training tasks; what is novel is only their pairing with the current instruction and scene (Task Perturbation) or their spatial placement (Position Perturbation). This ensures that the perturbations probe spatial and task generalization rather than confronting the policy with entirely unseen objects or regions. We run 10 rollouts per condition per task and report the average success rate across all 8 tasks.

\subsection{Grounding Source Details}
\label{supp:sec:subsec:real-world-grounding-source}
For all real-world experiments, we use Qwen3-VL-4B~\cite{qwen3vl} as our 2D grounding source. Unlike the simulation setup, where 2D target points are sampled from oracle segmentation masks, the real-world setup relies entirely on an off-the-shelf vision-language model: we apply a Qwen3-VL-4B directly with no fine-tuning. 

Given the task instruction and the front-view RGB image, we query the model with a single prompt that asks it to localize both sub-goal targets at once---the object to be picked up (role \texttt{target}) and the region or object where it should be placed (role \texttt{placement})---and to return their image-plane coordinates in a fixed JSON format. The exact prompt is shown below:
\begin{promptbox}
Task: "\{task\}"
\\

Based on this task, look at the image and identify:
\\

1. The object that needs to be picked up (role: "target")
\\
2. The location or object where it should be placed (role: "placement")
\\

Output ONLY a JSON array with exactly two entries, no extra text:

[

\qquad\{\{"point\_2d": [x, y], "label": "<object name>", "role": "target"\}\},

\qquad\{\{"point\_2d": [x, y], "label": "<object name>", "role": "placement"\}\}

]
\end{promptbox}

We parse the returned JSON and rescale the normalized coordinates (in $[0, 1000]$) to the input image resolution to obtain the two 2D points on the front-view image. Depending on the active sub-goal---reach-to-grasp or transport-to-place---the corresponding point (\texttt{target} or \texttt{placement}) is selected as the sub-goal target for the current time-step.

\subsection{Training Details}
\label{supp:sec:subsec:real-world-training-details}
\noindent\textbf{Dataset Details.}
Each frame in our collected real-world dataset records an $8$-D proprioceptive robot state, consisting of the end-effector pose---parameterized as the TCP position in the robot base frame together with a rotation vector (rotvec) for orientation ($6$-D)---and the measured gripper finger widths ($2$-D). The corresponding action is $7$-D and comprises a $6$-D end-effector command (position $+$ rotvec) and a $1$-D absolute gripper width command. Following~\cite{chi2024universal}, the end-effector action is expressed as a \emph{relative} transform with respect to the current end-effector pose, with the rotational component composed in $\mathrm{SO}(3)$.

\noindent\textbf{Training Details for GR00T-N1.6.}
We fine-tune GR00T-N1.6~\cite{gr00tn1_2025} on our collected real-world dataset. The model predicts an action chunk of $24$ future steps per forward pass. We use a global batch size of $256$ and train for $20{,}000$ steps using AdamW with a peak learning rate of $1{\times}10^{-4}$, a cosine decay schedule with a warmup ratio of $0.05$ ($\sim\!1{,}000$ steps), weight decay of $1{\times}10^{-5}$, and gradient clipping at a max norm of $1.0$. For visual augmentation, input images are first letter-boxed (padded with zeros on the shorter side) to a square, resized so that the shortest edge is $256$ px, randomly cropped along each side to $0.95$ of its length, and finally resized back to $256\times 256$. We further apply ColorJitter-style perturbations with brightness $0.3$, contrast $0.4$, saturation $0.5$, and hue $0.08$, while random rotation is disabled. All other hyperparameters follow the official configuration~\cite{gr00tn1_2025}.

\noindent\textbf{Training Details for $\pi_{0.5}$.}
We fine-tune $\pi_{0.5}$~\cite{pi05} on our collected real-world dataset. The model predicts an action chunk of $24$ future steps per forward pass. We use a global batch size of $256$ and train for $20{,}000$ steps using AdamW with a peak learning rate of $5{\times}10^{-5}$, a linear warmup schedule over the first $5{,}000$ steps and held constant thereafter, weight decay of $1{\times}10^{-10}$, and gradient clipping at a max norm of $1.0$. For visual augmentation, input images are resized (with padding) to $224{\times}224$ px; on the third-person view we apply a random crop covering $0.95$ of each spatial dimension, a bilinear resize back to $224{\times}224$ px, and a random rotation of $\pm 5^{\circ}$, while the wrist view receives no geometric augmentation. We further apply ColorJitter-style perturbations with brightness $0.3$, contrast $0.4$, and saturation $0.5$, while hue jitter is disabled. All other hyperparameters follow the official configuration~\cite{pi05}.

\subsection{Full Experimental Results}
\label{supp:sec:subsec:real-world-experimental-results-details}
\definecolor{mygray}{HTML}{EFEFEF}
\definecolor{myblue}{HTML}{EAF3FF}

\begin{table}[t]
\centering
\caption{Real-world In-Distribution evaluation results.}
\label{tab:real_in_distribution}
\resizebox{\linewidth}{!}{
\begin{tabular}{l cccccccc c}
\toprule
\textbf{Method} & Task 1 & Task 2 & Task 3 & Task 4 & Task 5 & Task 6 & Task 7 & Task 8 & \textbf{Avg.} \\
\midrule
$\pi_{0.5}$~\cite{pi05} (Baseline)         & 9/10  & 10/10  & 8/10  & 9/10  & 8/10 & 8/10 & 9/10  & 7/10 & 85.0 \\
\rowcolor{myblue}
$\pi_{0.5}$ + Ours              & 10/10 & 10/10 & 9/10 & 9/10  & 9/10 & 7/10 & 10/10 & 9/10 & 91.3 \\
\midrule
GR00T-N1.6~\cite{gr00tn1_2025} (Baseline)         & 7/10  & 8/10  & 8/10  & 7/10  & 5/10 & 7/10 & 7/10  & 6/10 & 68.8 \\
\rowcolor{myblue}
GR00T-N1.6 + Ours              & 10/10 & 10/10 & 10/10 & 8/10  & 9/10 & 9/10 & 10/10 & 8/10 & 92.5 \\
\midrule
GR00T-N1.6 + 2D via Text Prompt    & 9/10  & 9/10  & 8/10  & 8/10  & 9/10 & 9/10 & 10/10 & 8/10 & 87.5 \\
GR00T-N1.6 + 2D via Visual Prompt  & 9/10  & 8/10  & 8/10  & 10/10 & 9/10 & 8/10 & 10/10 & 9/10 & 88.8 \\
GR00T-N1.6 + 2D via AdaLN  & 10/10  & 9/10  & 7/10  & 9/10 & 9/10 & 8/10 & 9/10 & 9/10 & 87.5 \\
GR00T-N1.6 + 3D via Text Prompt & 8/10  & 9/10  & 9/10  & 8/10 & 8/10 & 9/10 & 9/10 & 9/10 & 86.3 \\
\bottomrule
\end{tabular}
}
\end{table}

\begin{table}[t]
\centering
\caption{Real-world Task Perturbation evaluation results.}
\label{tab:real_task_perturbation}
\resizebox{\linewidth}{!}{
\begin{tabular}{l cccccccc c}
\toprule
\textbf{Method} & Task 1 & Task 2 & Task 3 & Task 4 & Task 5 & Task 6 & Task 7 & Task 8 & \textbf{Avg.} \\
\midrule
$\pi_{0.5}$~\cite{pi05} (Baseline)         & 8/10  & 3/10  & 0/10  & 4/10  & 3/10 & 2/10 & 5/10  & 1/10 & 32.5 \\
\rowcolor{myblue}
$\pi_{0.5}$ + Ours              & 5/10 & 9/10 & 9/10 & 10/10  & 10/10 & 5/10 & 9/10 & 6/10 & 78.8 \\
\midrule
GR00T-N1.6~\cite{gr00tn1_2025} (Baseline)         & 0/10 & 3/10 & 0/10 & 0/10  & 0/10 & 0/10 & 0/10 & 0/10 & 3.8 \\
\rowcolor{myblue}
GR00T-N1.6 + Ours              & 9/10 & 8/10 & 0/10 & 10/10 & 9/10 & 7/10 & 3/10 & 5/10 & 63.8 \\
\midrule
GR00T-N1.6 + 2D via Text Prompt  & 0/10 & 4/10 & 0/10 & 0/10  & 0/10 & 0/10 & 0/10 & 0/10 & 5.0 \\
GR00T-N1.6 + 2D via Visual Prompt  & 0/10 & 3/10 & 0/10 & 1/10  & 0/10 & 1/10 & 0/10 & 0/10 & 6.3 \\
GR00T-N1.6 + 2D via AdaLN  & 3/10  & 5/10  & 0/10  & 2/10 & 3/10 & 2/10 & 0/10 & 0/10 & 18.8 \\
GR00T-N1.6 + 3D via Text Prompt & 2/10  & 0/10  & 0/10  & 0/10 & 0/10 & 1/10 & 0/10 & 1/10 & 5.0 \\
\bottomrule
\end{tabular}
}
\end{table}

\begin{table}[t]
\centering
\caption{Real-world Position Perturbation evaluation results.}
\label{tab:real_position_perturbation}
\resizebox{\linewidth}{!}{
\begin{tabular}{l cccccccc c}
\toprule
\textbf{Method} & Task 1 & Task 2 & Task 3 & Task 4 & Task 5 & Task 6 & Task 7 & Task 8 & \textbf{Avg.} \\
\midrule
$\pi_{0.5}$~\cite{pi05} (Baseline)         & 0/10  & 9/10  & 4/10  & 0/10  & 3/10 & 1/10 & 3/10  & 2/10 & 27.5 \\
\rowcolor{myblue}
$\pi_{0.5}$ + Ours              & 10/10 & 5/10 & 2/10 & 8/10  & 5/10 & 6/10 & 9/10 & 4/10 & 61.3 \\
\midrule
GR00T-N1.6~\cite{gr00tn1_2025} (Baseline)         & 2/10  & 0/10 & 0/10 & 0/10 & 0/10 & 0/10 & 0/10 & 1/10 & 3.8 \\
\rowcolor{myblue}
GR00T-N1.6 + Ours              & 10/10 & 5/10 & 9/10 & 9/10 & 0/10 & 3/10 & 9/10 & 8/10 & 66.3 \\
\midrule
GR00T-N1.6 + 2D via Text Prompt    & 3/10  & 0/10 & 0/10 & 0/10 & 0/10 & 0/10 & 0/10 & 3/10 & 7.5 \\
GR00T-N1.6 + 2D via Visual Prompt  & 1/10  & 0/10 & 0/10 & 0/10 & 0/10 & 1/10 & 0/10 & 2/10 & 5.0 \\
GR00T-N1.6 + 2D via AdaLN  & 4/10  & 0/10  & 2/10  & 0/10 & 0/10 & 1/10 & 5/10 & 1/10 & 16.3 \\
GR00T-N1.6 + 3D via Text Prompt & 2/10  & 0/10  & 0/10  & 0/10 & 0/10 & 1/10 & 0/10 & 1/10 & 5.0 \\
\bottomrule
\end{tabular}
}
\end{table}

We report numerical performance values for all evaluations in Sec.~\ref{sec:subsec:real-world-experimental-results}. Tab.~\ref{tab:real_in_distribution} presents the numerical results of In-Distribution evaluation, Tab.~\ref{tab:real_task_perturbation} presents the numerical results of Task Perturbation evaluation and Tab.~\ref{tab:real_position_perturbation} presents the numerical results of Position Perturbation evaluation.
\section{Qualitative Results and Failure Modes}\label{supp:sec:qualitative_results}
This section presents qualitative results in both simulation and real-world settings (Sec.~\ref{supp:subsec:qualitative_results}), followed by an analysis of failure modes in the real-world experiments (Sec.~\ref{supp:subsec:failure_modes}).

\subsection{Qualitative Results}\label{supp:subsec:qualitative_results}
\begin{figure}[t]
\centering
\includegraphics[width=0.9\linewidth]{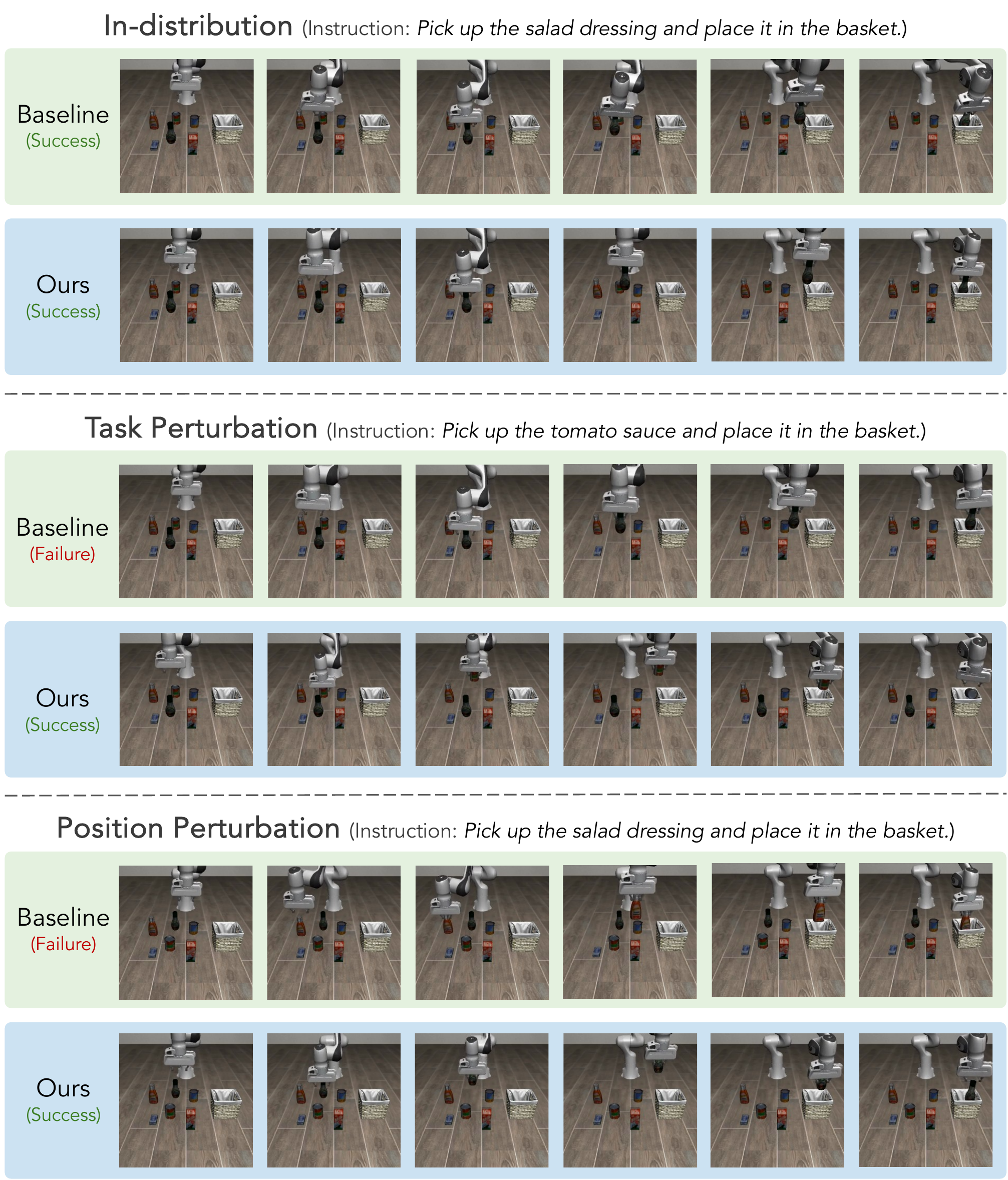}
\caption{\textbf{Qualitative results on LIBERO and LIBERO-PRO.}}
\label{fig:qualitative_results_sim}
\end{figure}
\begin{figure}[t]
\centering
\includegraphics[width=0.9\linewidth]{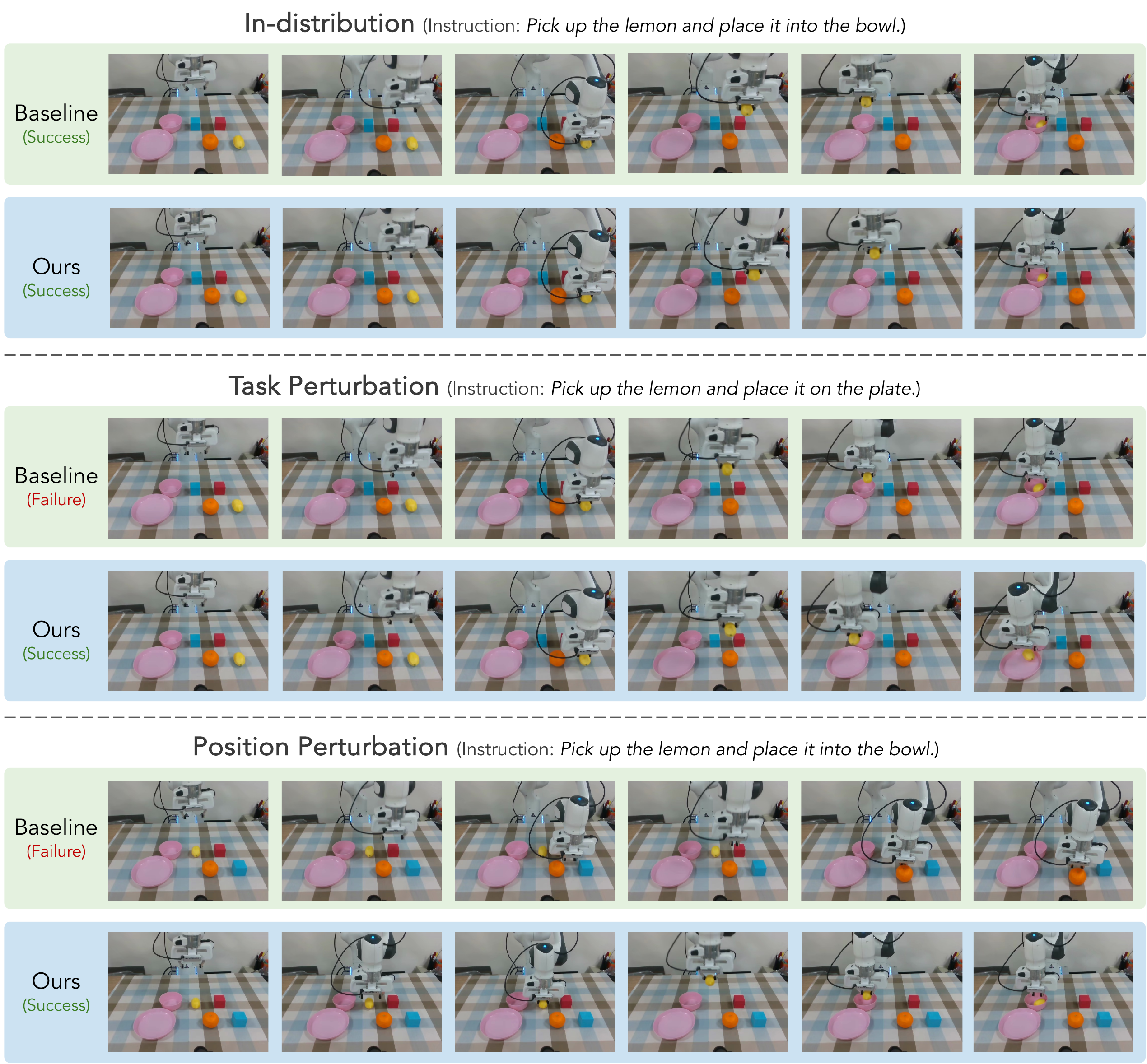}
\caption{\textbf{Qualitative results in the real-world setting.}}
\label{fig:qualitative_results_real_world}
\end{figure}
Fig.~\ref{fig:qualitative_results_sim} shows qualitative results on LIBERO and LIBERO-PRO, and Fig.~\ref{fig:qualitative_results_real_world} shows those in the real-world setting. In both simulation and the real world, the baseline succeeds under the In-distribution setting but fails under Task Perturbation and Position Perturbation, whereas our method remains successful under both perturbations. These observations are consistent with the quantitative results in Sec.~\ref{sec:libero_and_libero_pro} and Sec.~\ref{sec:subsec:real-world-experimental-results}. More qualitative results are available on our project page.

\subsection{Failure Modes in Real-World Experiments}\label{supp:subsec:failure_modes}
\begin{figure}[t]
\centering
\includegraphics[width=0.9\linewidth]{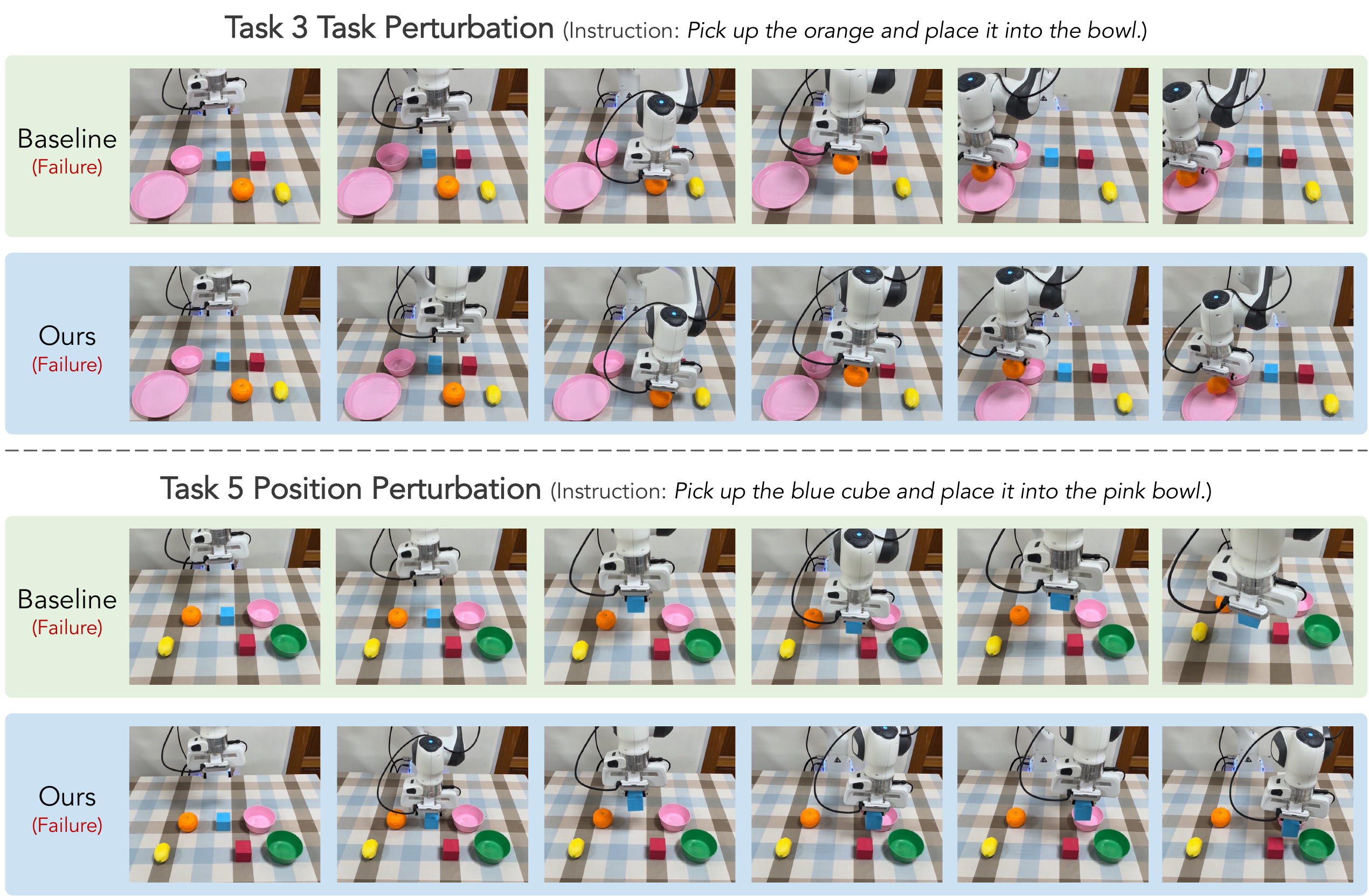}
\caption{\textbf{Failure modes in the real-world setting.}}
\label{fig:failure_cases_real_world}
\end{figure}
Although our method improves GR00T-N1.6~\cite{gr00tn1_2025} in the real-world experiments, it still achieves a 0/10 success rate in two settings: Task 3 under Task Perturbation and Task 5 under Position Perturbation. We analyze these two failure cases below, with representative rollouts of GR00T-N1.6 and our method shown in Fig.~\ref{fig:failure_cases_real_world}.

We first note that the 2D points predicted by Qwen3-VL-4B~\cite{qwen3vl} are correct in nearly all rollouts, suggesting that the failures do not stem from inaccurate grounding. Under Position Perturbation in Task 5, the policy attempts to adjust its trajectory toward the placement area but ultimately fails to complete the placement. Under Task Perturbation in Task 3, the policy places the object on the plate, i.e., the placement target seen during training, before reaching the bowl. We hypothesize that the larger size of the plate induces a broader 3D grounding region during training. As a result, at test time the policy interprets the grounding near the spatially adjacent bowl as falling within the learned plate region, leading to premature placement.
\end{document}